\documentclass[10pt,twocolumn,letterpaper]{article}

\usepackage[pagenumbers]{iccv} %

\input{preamble}

\usepackage{xcolor}       
\usepackage{enumitem}
\usepackage{amsmath}
\usepackage{amssymb}
\usepackage{mathrsfs}
\usepackage{mathtools}
\usepackage{amsthm}
\usepackage{wrapfig}
\usepackage[ruled]{algorithm2e}
\usepackage{float}
\usepackage{placeins}

\definecolor{iccvblue}{rgb}{0.21,0.49,0.74}
\usepackage[pagebackref,breaklinks,colorlinks,allcolors=iccvblue]{hyperref}

\title{FLASH\textmu: Fast Localizing And Sizing of Holographic Microparticles}
\author{First Author\\
Institution1\\
Institution1 address\\
{\tt\small firstauthor@i1.org}
\and
Second Author\\
Institution2\\
First line of institution2 address\\
{\tt\small secondauthor@i2.org}
}

\author{
  Ayush Paliwal\textsuperscript{1}, 
  Oliver Schlenczek\textsuperscript{1}, 
  Manuel Santos Pereira\textsuperscript{1}, 
  Birte Thiede\textsuperscript{1}, \\
  Katja Stieger\textsuperscript{1}, 
  Eberhard Bodenschatz\textsuperscript{1,3,4}, 
  Gholamhossein Bagheri\textsuperscript{1}, 
  Alexander Ecker\textsuperscript{2,1} 
  \vspace{5pt} \\  % Adds space between authors and affiliations
  \small  
  \textsuperscript{1}Max Planck Institute for Dynamics and Self-Organization, Göttingen, Germany \\
  \small
  \textsuperscript{2}Institute of Computer Science and Campus Institute Data Science, University of Göttingen, Germany \\
  \small
  \textsuperscript{3}Faculty of Physics, University of Göttingen, Friedrich-Hund-Platz 1, 37077 Göttingen, Germany\\
  \small
  \textsuperscript{4}Laboratory of Atomic and Solid State Physics, Cornell University, 523 Clark Hall, Ithaca, NY 14853, USA\\
  \footnotesize
  \texttt{\{ayush.paliwal, ecker\}@ds.mpg.de}, \;\texttt{Code:\href{https://github.com/ayushsvas/FlashMu.git}{https://github.com/ayushsvas/FlashMu}}
}

\newcommand{\fig}[1]{Fig.~\ref{#1}}
\newcommand{\tab}[1]{Table~\ref{#1}}
\newcommand{\eqn}[1]{Eq.\,(\ref{#1})}

\newcommand{\App}[1]{Appendix \ref{#1}}

\definecolor{lightblue3}{HTML}{cfe2f3}
\definecolor{lightred3}{HTML}{f4cccc}
\definecolor{lightyellow}{HTML}{FEF2D0}
\definecolor{lightgreen}{HTML}{DCE9D5}

\begin{document}
\maketitle
\begin{abstract}
Reconstructing the 3D location and size of microparticles from diffraction images -- holograms --  is a computationally expensive inverse problem that has traditionally been solved using physics-based reconstruction methods. More recently, researchers have used machine learning methods to speed up the process. However, for small particles in large sample volumes the performance of these methods falls short of standard physics-based reconstruction methods. Here we designed a two-stage neural network architecture, FLASHµ, to detect small particles (6--100\,µm) from holograms with large sample depths up to 20\,cm. Trained only on synthetic data with added physical noise, our method reliably detects particles of at least 9\,µm diameter in real holograms, comparable to the standard reconstruction-based approaches while operating on smaller crops, at quarter of the original resolution and providing roughly a 600-fold speedup. In addition to introducing a novel approach to a non-local object detection or signal demixing problem, our work could enable low-cost, real-time holographic imaging setups.
\end{abstract}

\section{Introduction}
\label{sec:intro}
\begin{figure*}[t!]
  \centering
  \begin{subfigure}{\linewidth}
    \includegraphics[width = \linewidth]{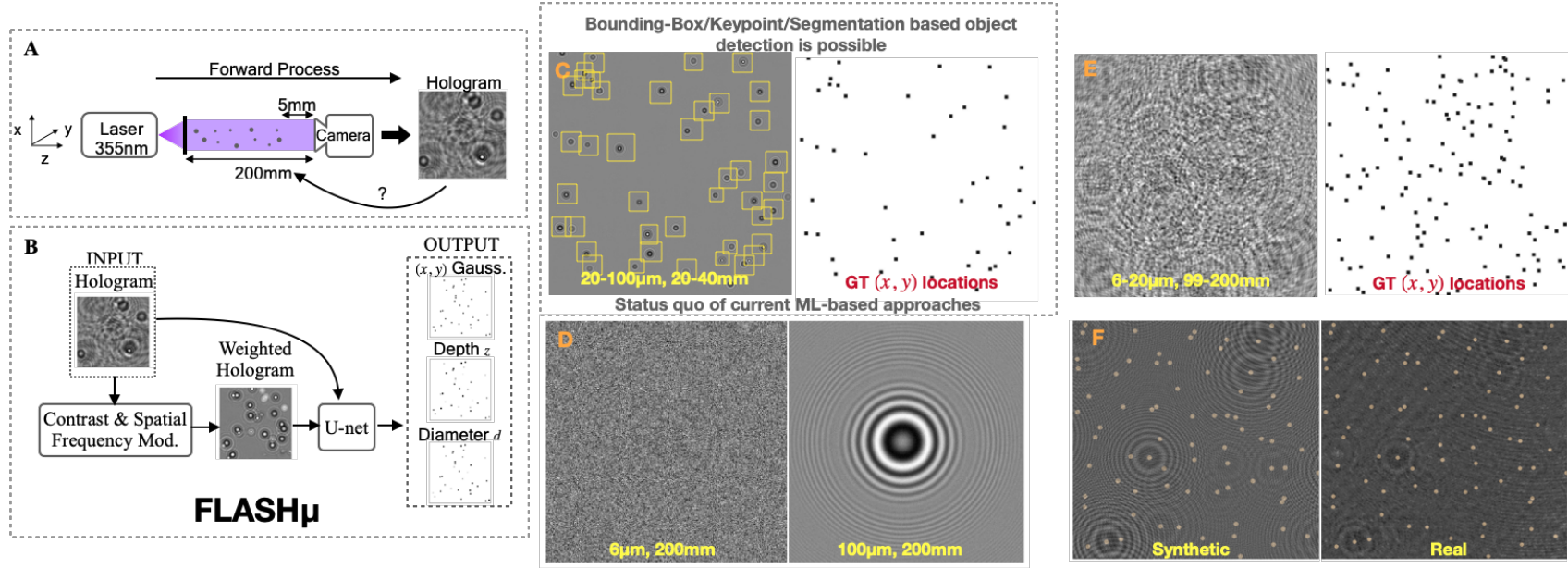}
  \end{subfigure}
\caption{Motivation and overview.
        \textbf{A.} Pictorial representation of the inverse problem.
        \textbf{B.} Schematic diagram of our approach: \textbf{FLASHµ}.
        \textbf{C.} Current ML-based approaches to hologram reconstruction rely on object detection, which limits their applicability to small sample depths (20-40\,mm) where fringe patterns are localized.
        \textbf{D.} The contrast of a particle is approximately proportional to its cross-sectional area, making small particles extremely hard to detect because of their low contrast.
        \textbf{E.} With smaller particles $\leq$ 20\,µm and increased sample depth, fringe patterns extend over almost the entire image and many particles' signals need to be demixed.
        \textbf{F.} Comparison between a synthetic hologram and a real hologram with ground truth (GT) overlaid. Real holograms have much lower signal-to-noise ratio.
    }
    
    \label{fig:teaser}
\end{figure*}

Holography is a versatile imaging technique that captures the diffraction patterns of objects on a 2D image, called hologram  (\fig{fig:teaser} A). During image acquisition, only the intensity variations are recorded, while the phases are averaged out \cite{ghatak2008optics}. In holography, the phase that encodes the 3D information about the scene can be reconstructed using the principles of diffraction of coherent light~\cite{gabor1948}.

Holography is typically used to characterize the shape, size and concentration of microparticles, i.e. small particles with diameters in the range of 0.1 to 100 micrometres. It has been extensively used for years in atmospheric sciences to study cloud particles \cite{thompson1974}, which are an important factor in understanding the Earth's climate and hydrological cycle \cite{doi:10.1126/science.1185138}.
Some of the uncertainties in weather and climate forecasting come from unresolved microphysical processes like the role of turbulence-driven collision-coalescence of cloud droplets in the formation of rain \cite{bertens2021experimental} or the spatial distribution of particles determining radiative transfer through clouds \cite{fugal_airborne_2004}. 
Resolving uncertainties in forecasting requires understanding of microphysical quantities, such as particle concentrations, sizes, and three-dimensional spatial distributions of individual droplets for a cloud, which requires probing large sample volumes $(\sim20\,$cm$^3$). In-situ setups like HOLODEC \cite{fugal_airborne_2004} or MPCK\textsuperscript{+} \cite{essd-13-4067-2021} record such holograms in abundance. However, analyzing a hologram is tedious~\cite{fugal2007insitu} and currently takes two to four orders of magnitude longer than acquiring it~\cite{fugal_airborne_2004, Schreck}. 
Real-time hologram reconstruction could be a game changer for applications in air quality monitoring, climate research and environmental management.

Fast machine learning approaches to hologram reconstruction based on standard object detection paradigms have been proposed \cite{Shimobaba:19, Shao:20, Chen, Zhang:22}, but they do not scale to large sample depths and real-world experimental conditions. They assume ideal localized particle images that can be delineated well by bounding boxes (\fig{fig:teaser}C), limiting the effective sample depth to a few centimeters. 
In deeper volumes, the overlapping fringe patterns generated by distant particles are distributed over the entire sensor, turning hologram reconstruction into a signal demixing rather than an object detection problem (\fig{fig:teaser}D). Small, distant particles are particularly challenging, because of their low contrast (note in \fig{fig:teaser}D--E how many of the particle's fringe patterns are not discernible). In addition, real holograms are noisy due to optics and other in-situ unknowns (\fig{fig:teaser}F), making synthetic-to-real generalization hard. 

Here we make the following contributions:
\begin{itemize}[nosep]
    \item We formulate holographic particle detection as a signal demixing and object localization problem, and propose FLASHµ -- a two-stage neural network architecture.  
    \item We propose dilations and Canonical Polyadic decomposition for spectral convolutions to scale to larger crop sizes.
    \item We introduce a simple yet effective data augmentation with empty holograms to improve sim-to-real transfer.
    \item Our approach can detect particles of at least 9\,µm (3\,px) diameter in real holograms, in volumes with a much higher sample depth than previous ML-based approaches.
    \item Operating on smaller image crops and lower resolution, our method is 600 times faster than standard approaches without significant loss of performance.  
\end{itemize}

\section{Background}
\label{sec:background}
In digital in-line holography, a reference (laser) beam $\phi_R\equiv\phi_R(x,y,z)$ illuminates a sample volume and diffracts (scatters) off the particles (\fig{fig:teaser}A). The diffracted light intensity is recorded on a pixel-based sensor (camera sensor). From this ``hologram'', a single 2D image of the recorded diffraction patterns, the spatial position and sizes of all particles within the sample volume can be reconstructed.

\paragraph{Forward Process.}
If $\phi_o $ is the diffracted field due to all individual objects, the resulting field $\phi $, at some distance $z$, is a linear superposition of the two, 
\begin{equation}
    \phi = \phi_R+\phi_o.
\end{equation}
The intensity (${I}$) reads
\begin{equation}
\label{eq:addition_complex}
\begin{aligned}
    {I}= |\phi|^2
     = |\phi_R|^2 + |\phi_o|^2 + {\phi^*_R\phi_o}+{\phi_R\phi^*_o},
\end{aligned}
\end{equation}
where $\phi^*$ denotes the complex conjugate of the field $\phi$. The last two terms in the above equation contain the phase which encodes the spatial information of the scene \cite{goodman2005introduction}. 
For a spherical particle with size $d_j=2r_j$, in the far-field approximation ($z_j\gg4r_j^2/\lambda$), located and observed at $(x_j,y_j,z_j)$ and $(x,y,z_o = 0)\equiv(x,y)$, respectively, diffraction is modelled using the Fraunhofer diffraction formula  \cite{thompson1974,goodman2005introduction},
    \begin{equation}\label{eq:fraunhofer}
        \phi^j_o(x,y) = \frac{r_j}{2i\rho_j}J_1\left(\frac{2\pi r_j\rho_j}{\lambda z_j}\right)\exp\left(\frac{i\pi\rho_j^2}{\lambda z_j}\right),
    \end{equation}
 where $\rho_j = \sqrt{(x-x_j)^2+(y-y_j)^2}$ is the distance between the particle center and a point in the camera plane ($z=0$) and $J_1$ is first-order Bessel function.   
For $N$ particles in the sample volume, we use
    \begin{equation}
    \label{eq:sum}
        I(k,l) = |\phi_R(k,l)+\sum_{j=1}^N\phi^j_o(k,l)|^2.
    \end{equation}
To synthetically model a hologram $\mathcal{H}$, we calculate the expected photon count at pixel $(k,l)$ of the square camera using \eqn{eq:fraunhofer} and \eqn{eq:sum}, added shot noise by sampling from a Poisson distribution and added Gaussian noise to simulate the sensor read noise \cite{sotthivirat2004penalized}, $\mathbf{b} \sim \mathcal{N}\!\left({0}, {\sigma}_{\mathrm{R}}\right)$,
\begin{equation}
\label{eq:measurement_model}
\mathcal{H}(k,l) \sim \text {Poisson}\Bigl(\!I(k,l)\!\Bigr)+{b}(k,l).
\end{equation}
$(\mathcal{H}, \left\{\left(x_j,y_j,z_j,r_j\right): j=1,2, \ldots, N\right\})$ forms labels for supervised learning. The task is to extract all the particles' information -- locations and sizes -- given a hologram $\mathcal{H}$. 

\section{Reconstruction methods}
\label{sec:prev_works}

\paragraph{Standard reconstruction.}
\label{sec:std_recon_method}

Also known as diffraction-based propagation, a hologram $\mathcal{H}$ is convolved with $g_z$, the Huygens-Fresnel kernel, to reconstruct the field (phase and amplitude) along the depth ($z$) dimension \cite{goodman2005introduction}, 
\begin{equation}
\label{eq:convolution_form}
    \mathcal{H}_z(\mathbf{r}) =\iint  d\mathbf{r}\,'\,\mathcal{H}_{z_o}(\mathbf{r}\,') g_{z-z_o}(\mathbf{r}-\mathbf{r}\,'),
\end{equation}
where $\mathbf{r}\equiv(x,y)$ and $\mathbf{r}\,'\equiv(x',y')$ are two-dimensional spatial vectors in camera plane $z_o=0$ and reconstructed plane $z$, respectively, with respect to the center of those planes. In practice, 
the `filtering' form of \eqn{eq:convolution_form} is used \cite{fugal_practical_2009}, which invokes the convolution theorem  
\begin{equation}
\label{eq:filtering}
\mathcal{H}_z(\mathbf{r})=
\mathcal{F}^{-1}\left\{G_{z-z_o}\cdot\mathcal{F}\{\mathcal{H}_{z_o}(\mathbf{r}\,')\}(\mathbf{k}\,')\right\},
\end{equation}
where $\mathcal{F}$, $\mathcal{F}^{-1}$ are Fourier and inverse Fourier transforms and 
\begin{equation}
G_{z-z_o}(\mathbf{k}) = \exp \left( \frac{2 \pi i (z-z_o)}{\lambda} \sqrt{1-\lambda^2 \mathbf{k}\cdot\mathbf{k}}\,\right). 
\end{equation}
with $\lambda$ being the wavelength and $i$ the imaginary unit. $G_{z-z_o}(\mathbf{k})$ is the Fourier representation of $g_{z-z_o}(\mathbf{r})$, and $\mathbf{k}$ is conjugate Fourier frequency vector to $\mathbf{r}$. 

In a reconstructed plane $z$, particles in focus appear as dark spots (as local minima; \fig{fig:toy}). To reliably identify particles from these reconstructions, further post-processing techniques are required. %
A brightness threshold alone is not selective enough to detect and size the particles as overlapping fringes in the reconstructed plane can be darker than the selected threshold giving rise to false detections and/or sizing biases \citep[see Sections 4.1.2 and 4.1.3 in][for details]{schlenczek2018airborne}. Therefore, one often uses edge filtering prior to thresholding~\cite{fugal_practical_2009, BAGHERI2023106102}. After thresholding, classification algorithms -- typically machine learning algorithms like SVMs, Decision Trees, CNNs -- trained on human-labeled data, are used along with some human supervision \cite{schlenczek2018airborne, amt-13-2219-2020,methodspaper} to label a dark spot as a particle/artifact.
\citet{Schreck} train neural networks to segment particles and extract their parameters from the reconstructed image. However, this approach does not circumvent the costly reconstruction process. 

With the standard reconstruction method, imaging small particles ($ d \geq 6$\,µm) far away (up tp 20\,cm) requires roughly a 20 megapixel sensor to record sufficient signal for reconstruction \cite{kreis_frequency_2002, fugal_practical_2009}, which becomes expensive as the computational complexity increases as $\mathcal{O}(K^2L \log K)$ for reconstructing $L$ slices of size $K\times K$. The operation is repeated $L$ times, once for each $z$ required to cover the entire depth of the volume. Typically, in cloud holography, one desires $100\,$µm depth resolution which amounts to reconstructing $\sim$2000 slices, all $20$ megapixels, which forms a substantial computational bottleneck.      

\begin{figure}[t]
  \centering
   \includegraphics[width=0.9\linewidth]{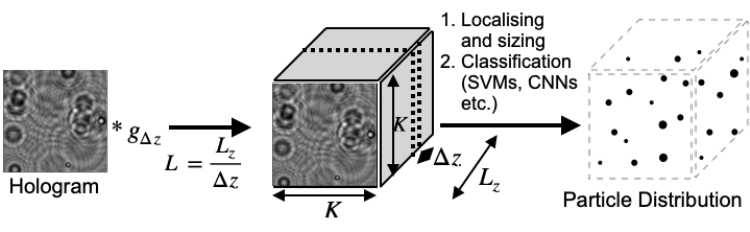}
   \caption{A schematic diagram describing the standard reconstruction-based method for particle extraction.}
   \label{fig:backpropdiff}
\end{figure}

\paragraph{Inverse-problem or optimization-based.}
With regularization, inverse-problem approaches have been extensively investigated to detect and localize particles in holograms \cite{fournier_inverse_2011}. Methods in the class of greedy algorithms have been developed \cite{Soulez:07, seifi_three-dimensional_2012}, iteratively estimating particle location and sizes with global detection and local optimization steps similar to matching pursuit \cite{Denis_2009}. Other compressed sensing methods \cite{NEEDELL2009301, Mallery:19}
which enforce sparsity and smoothness constraints with $l^1$ and TV regularization have been suggested. 
These methods are accurate and do overcome some of the sampling related problems of the standard reconstruction approach, but they are not suitable for processing voluminous data with thousands of particles because they are slow; typically even more computationally expensive than diffraction-based propagation methods. In addition, they require meticulous tuning of regularization parameters based on the particle number concentration, which is usually not known in advance for real-world holograms. GPU accelerated regularization methods performing FASTA \cite{Mallery:19}, differential holography \cite{partial3H} do bring up a speedup, but still do not avoid the volume reconstruction bottleneck.

\begin{figure*}[t!]
  \centering
  \begin{subfigure}{\linewidth}
    \includegraphics[width = \linewidth]{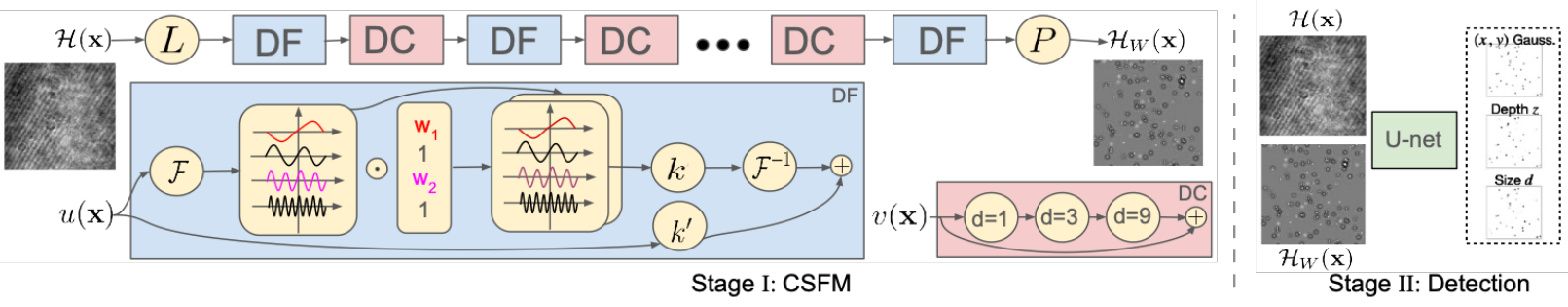}
  \end{subfigure}
  \caption{\textbf{Stage I: CSFM.} The hologram $\mathcal{H}$ is \circled[lightyellow][black][black]{\textit{L}}ifted to a multi-channel image, goes through several Dilated Fourier (DF) and Dilated Convolutional (DC) blocks, and \circled[lightyellow][black][black]{\textit{P}}rojected to Weighted Hologram $\mathcal{H}_w$. In the \colorbox{lightblue3}{DF} block, the input signal $u(\mathbf{x})$ is Fourier transformed $\mathcal{F}$, performs pointwise multiplication with the Fourier weights (w$_1$,w$_2$,\dots) convolved with kernel \circled[lightyellow][black][black]{$k$}, and taken back to pixel space with $\mathcal{F}^{-1}.$ The bottom skip-connection contains a convolutional kernel \circled[lightyellow][black][black]{$k'$}. The \colorbox{lightred3}{DC} inputs $v(\mathbf{x})$ which goes through three dilated convolutions with rates 1,3 and 9, and combines with a identity skip-connection. \textbf{Stage II: Detection.} The concatenated hologram, weighted-hologram $(\mathcal{H}, \mathcal{H}_w)$ pair is fed into the \colorbox{lightgreen}{U-net}. Localized fringes in $\mathcal{H}_w$ help the U-net to pinpoint (as $(x,y)$ Gaussian blobs) the particle with limited receptive field. Additionally, depth $z$ and size $d$ (diameter) are regressed as separate channels.}
  \label{fig:architecture}
\end{figure*}

\paragraph{Learning-based approaches.} 
Researchers have recently started applying machine learning techniques to directly estimate the particle parameters $(x_j,y_j,z_j,r_j)\in\mathbb{R}^4$ from a hologram $\mathcal{H}$ in a single inference step \cite{Shimobaba:19, Shao:20, Chen, Zhang:22}. 
While these methods do speed up the extraction-inference process, they have been evaluated only on holograms with small sample depths $\sim\,$2--3\,cm, which have relatively localized interference patterns that are straightforward to detect (\fig{fig:teaser}C). As we show below, vanilla end-to-end training of an object detector on holograms with larger sample depth on the order of 10\,cm fails to recover small, distant particles.

\section{Method: FLASHµ}
\label{sec:methodology}
The general idea of our approach is to first de-mix and localize the \textit{highly superposed and global} diffraction patterns from different particles such that a lightweight detection head can detect particles in 2D and regress their $z$ coordinate and size (\fig{fig:architecture}). This way we directly output a list of particle coordinates and sizes from the 2D hologram input without explicitly reconstructing the intensity-field volume.

\paragraph{Stage I: Contrast and Spatial Frequency  Modulator (CSFM).}
We observed that small and distant particles were regularly missed when larger particles were present, and hypothesized this was due to their low contrast in the image. This motivated a new intermediate representation -- ``weighted'' holograms $\mathcal{H}_w$: It keeps the $z$ information of the particles intact but has contrast and spatial frequencies redistributed in a way that all particles have similar contrast and have more localized fringes. We refer to the network generating this representation as Contrast and Spatial Frequency Modulator (CSFM). We obtain the weighted hologram $\mathcal{H}_w$ by setting the size of all particles to a fixed, large diameter ($d_w = 100$\,µm) and simulate the forward process with a lower wavelength ($\lambda_w = \lambda/8)$ in \eqn{eq:fraunhofer}. How to choose $d_w$ and $\lambda_w$ is discussed in \App{appendix:lam_w_hyperparas}. 

Because the fringe patterns of distant particles spread far across the image, working directly on the frequency space provides a good inductive bias; giving a global receptive field in the real space. 
Our network architecture for the CSFM step (\fig{fig:architecture}, left) is inspired by the Fourier Neural Operator (FNO) network \cite{kovachki2021neural, DBLP:journals/corr/abs-2010-08895} and Dilated Convolutional Neural Operators (DCNO) \cite{wang2020high, xu2024dilatedconvolutionneuraloperator}.
To achieve a good trade-off between field-of-view and resolution, the input is first downsampled by a $4\times4$ averaging kernel with stride 4, concatenated with positional embedding as $X-$channel and $Y-$channel, and is lifted to several latent channels using a linear layer. The core of the CSFM module consists of a series of dilated Fourier (DF; \fig{fig:architecture}, blue) and dilated convolutional (DC; \fig{fig:architecture}, red) blocks. 
The DF block is a modified version of he DCNO block~\cite{wang2020high, xu2024dilatedconvolutionneuraloperator}. As the number of parameters for the spectral convolution grows quadratically with image size, we introduce a stride $s$ (in practice we use $s=2$ most of the time). This means we include only every $s^\mathrm{th}$ frequency of the Fourier transformed signal in the pointwise multiplication with the Fourier kernel $w$; all other Fourier modes are left unchanged. The result is then concatenated with the input to provide context for the missing modes before being convolved with a kernel $k$. To reduce the number of parameters in the spectral convolution kernel $w$, we decompose it using canonical-polyadic (CP) decomposition, which generates a low-rank tensor. 
Then the processed signal is added to the skip connection, which contains a  $3\times 3$ convolution. The result is passed through a GeLU non-linearity. 

We alternate the dilated spectral block with dilated convolutional (DC) blocks. These DC blocks sequentially perform three convolutions with increasing dilation rates $\in {1, 3, 9}$. After each convolution, and addition after adding the skip connection, we apply a leaky ReLU. After sequential processing through these DF-DC blocks, a one-hidden-layer MLP projects the signal to the weighted hologram. 

This network is trained to minimize the squared error of the predicted hologram (network output) and the synthetic weighted hologram. All training details and hyperparameters of CSFM used are provided in \App{appendix:Training Details and Hyperparameters}.

\paragraph{Stage II: Detection.} 

Our detection head is a simple U-net \cite{DBLP:journals/corr/RonnebergerFB15}, with added modifications described in \cite{raonić2023convolutional} with residual blocks in skip connections and average pooling in place of max-pooling. A detailed description of the U-net can be found in \App{appendix:Training Details and Hyperparameters}.
The network thus outputs three feature maps (\fig{fig:architecture}, right) of size $K \times K$, with $K$ being the image width and height: (1) a heatmap $Y_{xy} \in [0,1]^{K \times K}$ of particle probabilities, (2) a $z$ map $Y_z \in [0,200\,\text{mm}]^{K \times K}$ and (3) a size map $Y_d \in [0, 100\,\text{µm}]^{K \times K}$ of particle diameters. The latter two are well defined only at particle locations and undefined (arbitrary values) elsewhere.
For training the network, we minimize $\mathcal{L} = \mathcal{L}_{xy} + \mathcal{L}^\text{Huber}_{z} + \mathcal{L}^\text{Huber}_{d}$.
The first term is the detection loss
\begin{equation}
\mathcal{L}_{xy}
 = \frac{1-\alpha}{K^2}\|Y-\hat{Y}\|_2^2 + \frac{\alpha}{K^2} \|\hat{Y}\|_{T V}^2,
\end{equation}
which consists of a squared error and a total variation (TV) term, which is the sum of spatial gradients across the image  and helps suppressing false positives \cite{Shao:20}. Here, $Y$ refers to the ground truth, generated by placing a small Gaussian window (SD 1\,px) centered on each particle location. $\hat{Y}$ denotes the model prediction; $\alpha = 10^{-4}$.

The second and third term are applied to the feature maps predicting $z$ and $d$. For these, we use the Huber loss \cite{10.1214/aoms/1177703732Huber} with $\delta = 5\times 10^{-4}$. \App{appendix:Training Details and Hyperparameters} details the exact values of hyperparameters used.

\paragraph{Hybrid holograms for sim-to-real transfer.}
A major challenge of training machine learning models on synthetic data is generalization to real-world data, because of the domain shift between idealized synthetic settings and all of the inaccuracies and noise sources in real-world data (\fig{fig:teaser}F). Such inaccuracies and noise in our case include inhomogeneities in the laser light, intensity fluctuations, a non-ideal light path, lens distortion and particles in the optical path outside the sampling volume, to name just a few. Such deviations from the idealized setting are challenging to account for in the forward model used to generate the synthetic data and and can vary from time to time and setup to setup. We therefore devised a simple but effective alternative method based on data augmentation, which we refer to as \textit{hybrid holograms}: We recorded a real-world ``empty'' hologram with no particles in the sample volume of the experimental setup. This experimental setup was different from the one we used for acquiring the data to test our model on real data. It had a different laser and different camera, thus making sure that we are not overfitting to the artifacts in the dataset we use for testing. We subtract the plane wave from this empty hologram and add it to the synthetic hologram:
\begin{equation}
    \mathcal{H}_{\text{hybrid}} = \mathcal{H}_{\text{synth}} + \mathcal{H}_{\text{empty}} - \mathcal{H}_{\text{plane}}
    \label{eq:hybrid}
\end{equation}
This is a coarse approximation, as the interactions are not additive adfter taking absolute values. However, it is still sufficient to force the network to learn to ignore any artifacts in the hologram. See \fig{fig:hybrid_holo_ex} in the appendix for an example.

\section{Experiments}
\label{sec:exp}

\paragraph{Training data.} 

We simulate the forward model with a coherent laser of wavelength 355\,nm using Eqs.~\eqref{eq:addition_complex}--\eqref{eq:measurement_model}, which includes realistic levels of sensor read noise (modeled as Gaussian) and Poisson shot noise. The resolution in $x$ and
$y$ is 3\,µm (= 1\,px). This corresponds to the sensor resolution typically used in real-world cloud holography \cite{fugal_airborne_2004, essd-13-4067-2021}. 

Our training data consists of three subsets: 

\textbf{(I)} A \textit{synthetic toy dataset} which is used as a minimal example to demonstrate that neural networks can reconstruct particle information beyond the resolution limit of standard reconstruction-based methods. It contains 50,000 holograms of size $512\times512$ pixels, each containing two particles with diameters 6\,µm and 100\,µm placed at 200\,mm from the camera. 

\textbf{(II)} A more \textit{realistic synthetic dataset} in which we simulate 13500 holograms for $4096\,\text{px} \times 4096\,\text{px}\times 195\,\text{mm}$ sample volume. These $4096\times4096$ holograms are then downsampled (by averaging) to $1024\times 1024$, going from a pixel pitch of 3\,µm to 12\,µm, from which crops of various sizes $\in\{128,256,384\}$ were taken. The density of particles was kept to be $\sim$70 particles/cm$^3$, which is in the range of typical cloud droplet particle concentrations. Particles were uniformly distributed in 3D and diameter $\in [6,100]\,$µm. 

Dataset \textbf{(III)} consists of \textit{hybrid holograms}, which are the same as \textbf{(II)}, but additionally augmented with empty backgrounds according to Eq.~\eqref{eq:hybrid}.  

\paragraph{Test data.}
To evaluate our models' performance, we test them on synthetic and real-world experimental data.

\textit{Synthetic.}
The synthetic test data has the same particle distribution as the synthetic training set \textbf{(II)}. Consisting of 4,000 holograms downsampled to $1024 \times 1024$. The locations and sizes of particles were drawn uniformly as in the training set \textbf{(II)}. 

\textit{CloudTarget: } For testing on real holograms, we use
the CloudTarget dataset, which consists of holograms of a set of photo masks and was developed to evaluate hologram processing~\cite{methodspaper}. The target consisted of a single quartz-fused silica photo mask with a refractive index (n) of 1.46, a thickness of 2.3\,mm, and 45 $\pm$ 8\,mm in width and height. The average particle density in the plane was about 400 particles/cm$^2$ and diameters ranged from 4--70\,µm. Size distribution in the target was skewed in a way that $\sim$80\% of the particles had diameters $< 20$\,µm, unlike the uniform size distribution of our training sets. The holographic measurements were obtained using the Advanced Max-Planck-CloudKites holographic system, which used a pulsed solid state laser ($\lambda = 355$\,nm) to illuminate the target and to create the holograms. The detector (Semi PYTHON25K Monochrome) had 5120 $\times$ 5120 pixels of 3\,µm effective width.  
The CloudTarget dataset consists of holograms with the target placed at low (50\,mm), medium (75\,mm, 99\,mm) and high depths (167.5\,mm, 192\, mm) relative to the image plane. The target was kept in motion laterally in the $xy$-plane. With this setup, the ground truth for the experiment is known, which is rare in microparticle holography.

\paragraph{Baselines.}

We compare to several baselines, grouped into reconstruction based, ML based and optimization based. 

\emph{Reconstruction based methods} start with 3D reconstruction as described in \cref{sec:std_recon_method}. After reconstruction and intensity thresholding, potential particles need to be classified as artifacts or real particles. For this classification, we trained two sets of three SVMs, referred to in the following as SVM1, SVM2 and SVM3. The first set of SVMs was trained on a portion of synthetic test data. The second set was trained on a portion of ground truth labels of CloudTarget. To generate labels for the particle--artifact classification, trained experts manually classified reconstructed candidate crops as either sure particles, unsure objects or sure artifacts. To train SVM1 we used only sure particles and sure artifacts for the two classes. For SVM2 and SVM3 we used all three classes. SVM2 optimized for recall and treated unsure objects like particles, while SVM3 optimized for precision and treated them as artifacts for the final predictions. SVMs (denoted as Recon-SVM) form our strongest baseline since the classifier is directly trained on real-world data. The second baseline in this group is the recent  standard reconstruction-based CNN method \cite{methodspaper} (Recon-CNN) similar to \cite{Schreck}. Here the model is trained on manually labeled real-world cloud holograms. We compare our method with this baseline on CloudTarget. A cautionary note: The glass target in the CloudTarget experiment produced artifacts, referred to as ``ghost layers''~\cite{methodspaper}, presumably due to internal reflections within the glass target, which leads to slightly inflated false detection rates by FLASHµ and all other baselines. As a result, the reported precision levels on CloudTarget are lower than would be expected under real-world conditions where no glass sheets are within the sampling volume. Also note that the sizing threshold of Recon-SVM and Recon-CNN has been fine-tuned on CloudTarget, putting these baselines at an advantage over our method.    

In the \emph{ML based group}, we compare our method to three other strong architectures developed for dense prediction problems like segmentation, depth estimation or keypoint detection. 1) U-net-based detector~ \cite{DBLP:journals/corr/RonnebergerFB15}, which is slightly modified following \citet{raonić2023convolutional}; 2) DeepLabV3 \cite{chen2017rethinkingatrousconvolutionsemantic} with ResNet-101 backbone; and 3) UperNet with a ConvNeXt backbone \cite{liu2022convnet2020s}. These architectures are trained on the same objective as FLASHµ's detection head, similar to \citet{Shimobaba:19, Shao:20}, but without the CSFM. Note that bounding-box based object detection is not possible for these holograms as small, distant particles can spread across the entire image (\fig{fig:teaser}). Hence we cannot use object detection based baselines.

In the \emph{optimization based group}, we experimented with the approach of \citet{ni_chen_diffHolo}($\partial H^3$). However, the convergence and, hence, the performance of this (and other) optimization based methods is highly sensitive to the regularization coefficients, which require fine-tuning \textit{per hologram} as they depend on particle density, particle radius, and overall volume size \cite{ni_chen_diffHolo}. We therefore consider this class of methods not sufficiently general and provide only the inference time for comparison with our method and other baselines.

\paragraph{Evaluation.}
To evaluate our models, we compute precision--recall curves for varying detection thresholds. Particle detections are obtained by finding all local maxima in the predicted probability map of the detection network which are above the threshold. Detections are counted as hits if their $(x,y)$ position is within a $7\times7$ window of a ground truth particle and the regressed $z$ coordinate is within 10\,mm of the real value. Otherwise a detection is counted as a false positive. All ground truth particles not matched by a detection are counted as misses.
To measure the quality of our predictions for the depth ($z$) and size ($d$) of detected particles within the sample, we report the median and interquartile range of the absolute error. 

\section{Results}
\label{sec:results}

\paragraph{Reconstruction beyond the resolution limit.} 

We start by demonstrating how small particles far away cannot be reconstructed using the standard reconstruction method (described in Section \ref{sec:std_recon_method}) due to the diffraction limit. The diffraction limit, also known as the Rayleigh criterion \cite{doi:10.1080/14786447908639684Rayleigh}, is the maximal distance at which one can resolve a particle using standard reconstruction methods.
For a circular particle with 6\,µm diameter and a camera with 3\,µm pixel pitch and $512\times 512$ pixel sensor size, this is roughly $z_{\text{lim}} \approx D_{\text{det}}\cdot d / 2.44\lambda \approx 11\,\text{mm}$. 

\begin{figure}[t]
  \centering
   \includegraphics[width=0.85\linewidth]{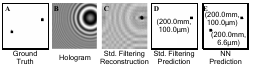}
   \caption{Toy example illustrating how NNs can overcome the limits of reconstruction-based methods in terms of sensor size and resolution. 
    \textbf{A:}~Two particles (6\,µm and 100\,µm), placed 200\,mm from the camera. 
    \textbf{B:}~Resulting $512\times 512$ hologram. 
    \textbf{C:}~Standard reconstruction at $z = 200$\,mm. 
    \textbf{D:}~Only $100$\,µm particle is detected from the reconstruction.
    \textbf{E}:~U-net (shown here) and FLASHµ trained on toy dataset (\textbf{I}) reliably detect both particles.}
   \label{fig:toy}
\end{figure}

\begin{figure}[b]
  \centering
   \includegraphics[width=\linewidth]{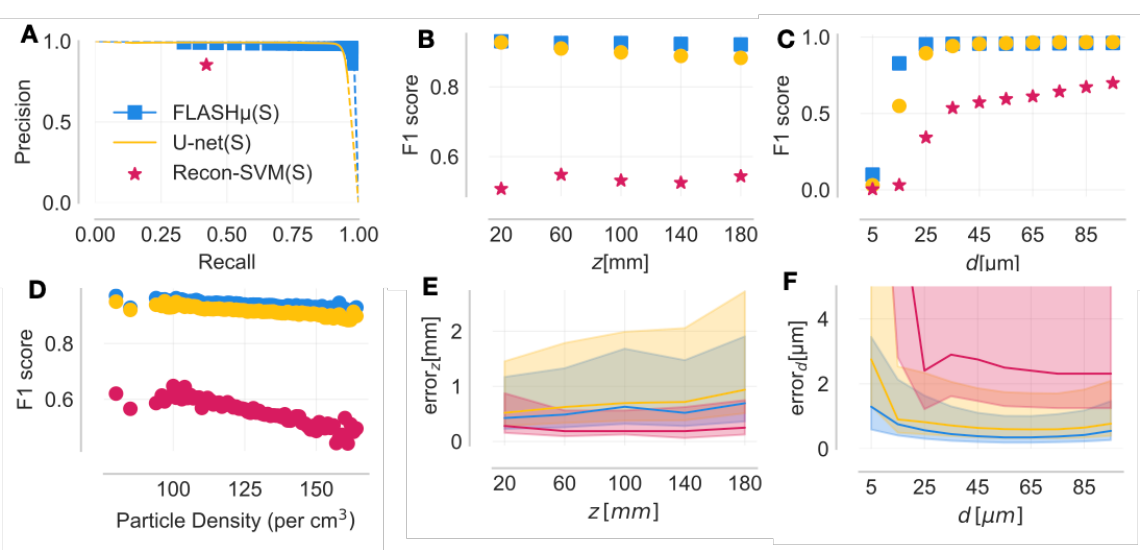}
   \caption{Quantitative evaluation of FLASHµ and baselines (U-net and SVMs) on 384$\times 384$ (S)ynthetic holograms. 
    \textbf{A:}~Precision over recall for the synthetic test dataset. 
    \textbf{B:} F1 score as function of distance from camera $z$.
    \textbf{C:} F1 score as function of particle diameter $d$.
    \textbf{D:} F1 score as function of particle number density.
    \textbf{E,\,F:} Comparison of median and interquartile of absolute error in $z$ with respect to $z$ (\textbf{E}) and $d$ (diameter) (\textbf{F}).}
   \label{fig:results_synthetic}
\end{figure}

\begin{figure*}[t!]
  \centering
  \begin{subfigure}{0.22\linewidth}
    \includegraphics[width = \linewidth]{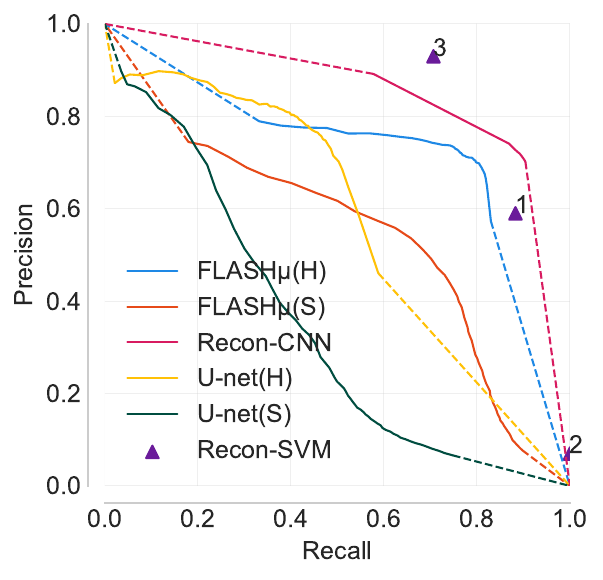}
    \caption{Precision-Recall curves of different methods on CloudTarget, trained using synthetic (S) or hybrid (H) holograms. }
    \label{fig:prec_rec}
  \end{subfigure}
  \hfill
  \begin{subfigure}{0.77\linewidth}
    \centering
    {\small
    \begin{tabular}{c|c|c|c|c|c|c}
        Approach & Method &  Precision $\uparrow$ & Recall $\uparrow$ & F1 $\uparrow$ & \#P(GB) $\downarrow$ & Inference time (sec) $\downarrow$ \\
        \hline
        ML, 1-stage & DeepLabV3 & 0.28 & 0.34 & 0.30 & \textcolor{black}{0.22} & 0.058 $\pm$ 0.001\\
        ML, 1-stage & ConvNeXt & 0.69 & 0.47 & 0.56 & \textcolor{black}{0.22}  & 0.088 $\pm$ 0.002 \\
        ML, 1-stage & U-net  &  0.74 & 0.55 & 0.63 & 0.22 & 0.036 $\pm$ 0.001\\
        ML, 2-stage & \textbf{FLASHµ}  & 0.71 & 0.80 & 0.75 & 0.25 &  0.494 $\pm$ 0.014\\
        \hline
        Reconstruct. & SVM  & 0.91 & 0.70 & 0.77 &  $\sim 0$ & $> 300_{gpu}, > 600_{cpu}$\\
        Reconstruct. & CNN  & 0.76 & 0.85 & 0.79&  $\sim 0$ & $> 300_{gpu}, > 600_{cpu}$\\
        \hline
        Optimization & $\partial H^3$ & - & - & - & - & $\ggg 300_{gpu}, \ggg 600_{cpu}$\\
    \end{tabular}}
    \label{tab:comparison_baselines_inference_speed}
    \caption{Overview of performance, including detection metrics (precision, recall at peak F1 scores), model size (\#P = number of parameters) and inference speeds in seconds per $5120\times 5120$ hologram.
    Inference times (on a single 40GB A100 GPU) of reconstruction and optimization based methods are lower bounds of complete processing (hologram-to-list of particles) as it does not include post-processing steps. For ML-based approaches, it is hologram-to-list of particles.}
  \end{subfigure}
  \caption{Comparison with baselines and the effect of hybrid holograms on CloudTarget dataset. Note: Precision levels are a conservative estimate of what is expected in reality, since the CloudTarget's glass layer generates reflections that cause false detections~\cite{methodspaper}.}
  \label{fig:comparison_baselines}
\end{figure*}

Consider a small toy example (\fig{fig:toy}) with two particles at 200\,mm depth (way beyond $11$\,mm), the small one 6\,µm in diameter, the larger one 100\,µm. The standard reconstruction method cannot bring the 6\,µm particle in focus in the presence of the larger one (\fig{fig:toy}C--D). In contrast, our neural networks (both the baseline U-net and FLASHµ) are able to detect and size both particles correctly (\fig{fig:toy}E). This toy example shows that detection of small and distant particles is possible beyond the diffraction-based resolution limit with neural networks. It provides a rationale for the hypothesis that holography can be done on smaller and cheaper scales using neural networks than existing classical techniques.

\paragraph{FLASHµ outperforms standard reconstruction on synthetic holograms.}

Next we go beyond the two-particle case  and provide quantitative results for the more general case of synthetic holograms with multiple particles of varying size distributed randomly in the probe volume. 
FLASHµ outperforms the reconstruction-based SVM baseline by a large margin and also shows a clear, albeit smaller, improvement over the U-net baseline (\fig{fig:results_synthetic}).
The improvements in the detection rate (recall) and precision are evident (\fig{fig:results_synthetic}A) across all particle depths (\fig{fig:results_synthetic}B), sizes (\fig{fig:results_synthetic}C) and particle densities (\fig{fig:results_synthetic}D). The advantage of FLASHµ over the U-net baseline is evident in regressing $z$ (\fig{fig:results_synthetic}E), better detecting small sizes  (\fig{fig:results_synthetic}C, F) and for more densely packed volumes (\fig{fig:results_synthetic}D).

\begin{figure}[b]
    \centering
    \vspace{-3pt}
    \includegraphics[width=\linewidth]{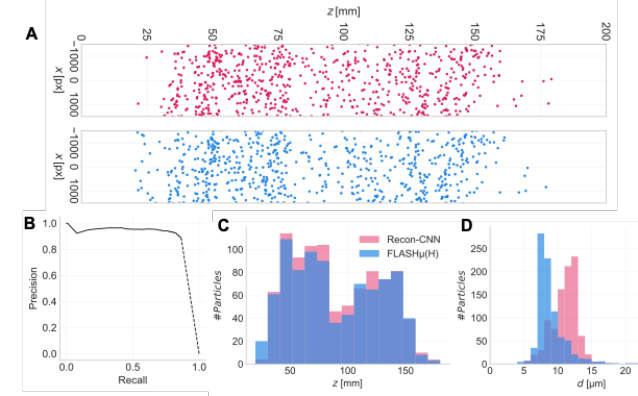}
    \caption{Evaluation of FLASHµ on real medium-density cloud hologram, treating Recon-CNN as ground truth. \textbf{A:} Visual of the ($x,z$) predictions. \textbf{B:} Precision-recall of FLASHµ with respect to Recon-CNN.
    Predicted histograms of depth (\textbf{C}) and diameter (\textbf{D}) from the two methods.\vspace{-3pt} }
    \label{fig:medium_cloud}
\end{figure}

\paragraph{FLASHµ achieves sim-to-real transfer.} 

We now ask how our approach generalizes to real data, using the CloudTarget test set. Models trained on synthetic data often fail to generalize to real-world data due to domain shifts. This is also true for FLASHµ and the U-net baseline when trained only on synthetic data (\fig{fig:comparison_baselines}A, orange and black). However, when training on hybrid holograms -- generated by augmenting synthetic ones with empty holograms -- generalization improves greatly (\fig{fig:comparison_baselines}A, yellow and blue) and approaches the performance of the standard reconstruction method (\fig{fig:comparison_baselines}A, pink and triangles) despite operating on a lower resolution and smaller crops. 

FLASHµ is the only ML-based approach that achieves competitive performance with reconstruction based methods (\fig{fig:comparison_baselines}B). Other strong image-to-image architectures did not achieve comparable performance, showing that our two-stage approach using CSFM is indeed effective (\fig{fig:comparison_baselines}B). This additional stage adds only 30\,MB of parameters but increases the computational cost by roughly one order of magnitude due to the spectral convolutions. It is still more than two orders of magnitude faster, though, than the standard reconstruction method (\fig{fig:comparison_baselines}B, last column). Note that all methods were benchmarked on equal grounds, using a single 40GB A100 GPU.

We next asked how FLASHµ performs across the depth of the sample volume and for different particle sizes, and found that it achieves similar or even better recall than the standard reconstruction method for particles greater than 9\,µm (\App{appendix:error_size_z_hist}, \fig{fig:error_plots}, first row). The sizing errors are comparable (\fig{fig:error_plots}, second row) despite the reconstruction based method being tuned specifically to CloudTarget. Regressing the $z$ coordinate also works decently, albeit with significantly larger error than the standard reconstruction method (\fig{fig:error_plots}, third row).

Finally, we give a teaser of how our method performs on real-world holograms measured in real clouds (\fig{fig:medium_cloud}). Compared against the carefully tested reconstruction based method~ \cite{methodspaper}, our method performs decently for sparse to medium-dense clouds ($\sim$70 particles/cm$^3$), achieving around 90\% in terms precision-recall (\fig{fig:medium_cloud}B) and predicting similar particle distributions across $z$ (\fig{fig:medium_cloud}C). Note the (moderate) bias in sizing (\fig{fig:medium_cloud}D). We provide two more examples of low- and high-density clouds in \App{appendix:cloud_holograms}.

\paragraph{How does the method scale?}

Having established that our method works, we consider some of its design choices. First, as expected, performance improves with increased crop size (\tab{tab:ablation_crop_sz}). 

\begin{table}[h]
    \centering
    \small
    \begin{tabular}{c|r|r|r}
        Crop size & P $\uparrow$ & R $\uparrow$ & F1 $\uparrow$ \\
        \hline 
        128 &  0.73 & 0.48 & 0.58   \\
        256 & 0.72 & 0.59 & 0.65 \\
        \textbf{384} & \textbf{0.70} & \textbf{0.73} & \textbf{0.72}      
    \end{tabular}%
    \caption{Effect of increasing crop size on (P)recision, (R)ecall and peak F1 score, fixing the FLASHµ architecture.}
    \label{tab:ablation_crop_sz}
\end{table}

Second, the range of frequencies covered by the spectral convolution matters. As the number of parameters grows quadratically with the number of frequencies included, we first started by considering only a small number of low frequencies but this hurts performance (\tab{tab:ablation_ker_sz}; 48--96). We therefore opted for the approach of covering the full frequency range but with a stride, which resulted in the best trade-off between model size and performance (\tab{tab:ablation_ker_sz}).

\begin{table}[h]
    \centering
    \small
    \begin{tabular}{c|c|r|r|r|r|r}
        Kernel size & Stride & P $\uparrow$ & R $\uparrow$ & F1 $\uparrow$ & \#P $\downarrow$ & T $\downarrow$ \\
        \hline 
        48 & 1 &  0.68& 0.72& 0.69 & 0.58 & 0.36 \\
        64 & 1 & 0.70& 0.73& 0.72 & 1.02 & 0.36   \\
        96 & 1 &  0.69& 0.72& 0.70 &  2.27 & 0.36 \\
        192 & 1 & - & - &   - & 8.79 & -\\
        192 & 2 & 0.70& 0.77& 0.73 & 2.28 &0.47\\
        384 & 1 & - & - & - & 35.79 & - \\
        \textbf{384} & \textbf{4} & \textbf{0.69} & \textbf{0.80} & \textbf{0.74} & \textbf{2.29} &  \textbf{0.71} \\
    \end{tabular}
    \caption{Effect of increasing the Fourier kernel size on (P)recision, (R)ecall and peak F1 score at fixed crop size of 384. T: inference time in seconds per $5120 \times 5120$ hologram.  \#P: number of FP32 parameters (GB) of CSFM. U-net is 0.22\,GB. Rows left blank could not be trained due to memory limits.  }
    \label{tab:ablation_ker_sz}
\end{table}

Because the number of parameters in the spectral convolutions is still very large, we reduce it further by performing the spectral convolution depthwise, i.\,e. per channel, and using canonical-polyadic (CP) decomposition, a low-rank decomposition of the 3D weight tensor. Taken together, this lets us scale to 384 $\times$ 384 resolution with a stride of only 2, which further improves performance while bringing down the number of parameters to just 30\,MB (\tab{tab:ablation_cp_sep}).

\begin{table}[h]
    \centering
    \small
    \begin{tabular}{c|c|c|c|r|r|r|r|r}
        Cs & Str & Rank & D & P $\uparrow$ & R $\uparrow$ & F1 $\uparrow$ & \#P $\downarrow$ & T $\downarrow$ \\
        \hline 
         384 & 2 &Full & N & - & - & - &          8.80 & - \\
         384 & 2 &Full & Y & 0.72 & 0.79 & 0.75 & 2.33 & 0.50\\
         384 & 2 &2048 & Y & 0.72 & 0.79 & 0.75 & 0.04 & 0.57\\
         \textbf{384} & \textbf{2} &\textbf{16}  & \textbf{Y} & \textbf{0.71} & \textbf{0.80} & \textbf{0.75} & \textbf{0.03} & \textbf{0.49}
    \end{tabular}
    \caption{Effect of CP decomposition and (D)epthwise multiplication in  Fourier blocks. Cs: Crop Size. Str: stride. Rank: Kruskal rank. T: inference time in seconds per $5120
    \times 5120$ hologram. \#P: number of FP32 parameters (GB) of CSFM. U-net is 0.22\,GB. 
    }
    \label{tab:ablation_cp_sep}
\end{table}

In \App{appendix:ablations}, we also show how removing the DC-blocks degrades the performance (\tab{tab:ablation_add_dc}) and that interpolation is important (\tab{tab:ablation_interp}). Further ablations on number of training samples (\tab{tab:ablation_training_samples}), number of layers (\tab{tab:ablation_n_layers}) and width (\tab{tab:ablation_width}) are shown there as well.

\section{Conclusion}
\label{sec:conclude}
FLASHµ is the first end-to-end learning-based approach for holographic reconstruction that achieves performance comparable to standard reconstruction-based methods in settings with large sample volumes where traditional bounding-box, keypoint or segmentation based ML approaches fail. By introducing CSFM, our method transforms global fringe patterns into local ones and equalizes the contrast of differently sized particles, which enables detection viable with a simple U-net. Our approach of dilating the spectral convolution and using a low-rank representation of the weights could also be useful more broadly where FNO-type networks are being used.

The impact of being able to do particle detection from holograms more than two orders of magnitude faster could be substantial. Researchers have collected on millions of cloud holograms, many of which can now be processed on a much more feasible time horizon.
In addition, FLASHµ reliably operates at smaller scales -- smaller crop size, coarser pixel resolution and small model size, which could enable the design of low-cost small-scale holographic imaging setups with applications in real-time 3D imaging of particle flows like airborne transmission of infectious diseases, indoor monitoring of pollutants, or detecting water vapour in the atmosphere of other planets in space.

\section{Limitations}
\label{sec:limits}

We would like to discuss a few shortcomings of our method: First, the $z$ resolution could be improved. The current resolution will preclude certain analyses where high $z$ resolution is necessary.
Second, the forward model assumes spherical particles. Our results are promising, but it is unclear how robust our method is to deviations from this assumption, e.\,g. when measuring ice clouds or other non-spherical particles.

{
    \small
    \bibliographystyle{ieeenat_fullname}
    \bibliography{biblio}

\begin{thebibliography}{39}
\providecommand{\natexlab}[1]{#1}
\providecommand{\url}[1]{\texttt{#1}}
\expandafter\ifx\csname urlstyle\endcsname\relax
  \providecommand{\doi}[1]{doi: #1}\else
  \providecommand{\doi}{doi: \begingroup \urlstyle{rm}\Url}\fi

\bibitem[Bagheri et~al.(2023)Bagheri, Schlenczek, Turco, Thiede, Stieger,
  Kosub, Clauberg, Pöhlker, Pöhlker, Moláček, Scheithauer, and
  Bodenschatz]{BAGHERI2023106102}
Gholamhossein Bagheri, Oliver Schlenczek, Laura Turco, Birte Thiede, Katja
  Stieger, Jana~M. Kosub, Sigrid Clauberg, Mira~L. Pöhlker, Christopher
  Pöhlker, Jan Moláček, Simone Scheithauer, and Eberhard Bodenschatz.
\newblock Size, concentration, and origin of human exhaled particles and their
  dependence on human factors with implications on infection transmission.
\newblock \emph{Journal of Aerosol Science}, 168:\penalty0 106102, 2023.

\bibitem[Bertens(2021)]{bertens2021experimental}
Guus Bertens.
\newblock \emph{Experimental investigation of cloud droplet dynamics at the
  research station Schneefernerhaus}.
\newblock PhD thesis, Georg-August University School of Science (GAUSS)
  G{\"o}ttingen, 2021.

\bibitem[Bodenschatz et~al.(2010)Bodenschatz, Malinowski, Shaw, and
  Stratmann]{doi:10.1126/science.1185138}
E. Bodenschatz, S.~P. Malinowski, R.~A. Shaw, and F. Stratmann.
\newblock Can we understand clouds without turbulence?
\newblock \emph{Science}, 327\penalty0 (5968):\penalty0 970--971, 2010.

\bibitem[Chen et~al.(2017)Chen, Papandreou, Schroff, and
  Adam]{chen2017rethinkingatrousconvolutionsemantic}
Liang-Chieh Chen, George Papandreou, Florian Schroff, and Hartwig Adam.
\newblock Rethinking atrous convolution for semantic image segmentation, 2017.

\bibitem[Chen et~al.(2021)Chen, Wang, and Heidrich]{Chen}
Ni Chen, Congli Wang, and Wolfgang Heidrich.
\newblock Holographic 3d particle imaging with model-based deep network.
\newblock \emph{IEEE Transactions on Computational Imaging}, 7:\penalty0
  288--296, 2021.

\bibitem[Denis et~al.(2009)Denis, Lorenz, and Trede]{Denis_2009}
L Denis, D~A Lorenz, and D Trede.
\newblock Greedy solution of ill-posed problems: error bounds and exact
  inversion.
\newblock \emph{Inverse Problems}, 25\penalty0 (11):\penalty0 115017, 2009.

\bibitem[Fournier et~al.(2011)Fournier, Denis, Thiebaut, Fournel, and
  Seifi]{fournier_inverse_2011}
Corinne Fournier, Loic Denis, Eric Thiebaut, Thierry Fournel, and Mozhdeh
  Seifi.
\newblock {Inverse problem approaches for digital hologram reconstruction}.
\newblock In \emph{Three-Dimensional Imaging, Visualization, and Display 2011},
  page 80430S. International Society for Optics and Photonics, SPIE, 2011.

\bibitem[Fugal(2007)]{fugal2007insitu}
Jacob~P. Fugal.
\newblock \emph{In-situ measurement and characterization of cloud particles
  using digital in-line holography}.
\newblock Dissertation, Michigan Technological University, 2007.

\bibitem[Fugal et~al.(2004)Fugal, Shaw, Saw, and Sergeyev]{fugal_airborne_2004}
Jacob~P. Fugal, Raymond~A. Shaw, Ewe~Wei Saw, and Aleksandr~V. Sergeyev.
\newblock Airborne digital holographic system for cloud particle measurements.
\newblock 43\penalty0 (32):\penalty0 5987, 2004.

\bibitem[Fugal et~al.(2009)Fugal, Schulz, and Shaw]{fugal_practical_2009}
Jacob~P Fugal, Timothy~J Schulz, and Raymond~A Shaw.
\newblock Practical methods for automated reconstruction and characterization
  of particles in digital in-line holograms.
\newblock \emph{Measurement Science and Technology}, 20\penalty0 (7):\penalty0
  075501, 2009.

\bibitem[Gabor(1948)]{gabor1948}
D. Gabor.
\newblock A new microscopic principle.
\newblock 161:\penalty0 777–78, 1948.

\bibitem[Ghatak(2008)]{ghatak2008optics}
A. Ghatak.
\newblock \emph{Optics}.
\newblock Tata McGraw-Hill, 2008.

\bibitem[Goodman(2005)]{goodman2005introduction}
J.W. Goodman.
\newblock \emph{Introduction to Fourier Optics}.
\newblock W. H. Freeman, 2005.

\bibitem[Huber(1964)]{10.1214/aoms/1177703732Huber}
Peter~J. Huber.
\newblock {Robust Estimation of a Location Parameter}.
\newblock \emph{The Annals of Mathematical Statistics}, 35\penalty0
  (1):\penalty0 73 -- 101, 1964.

\bibitem[Kovachki et~al.(2021)Kovachki, Li, Liu, Azizzadenesheli, Bhattacharya,
  Stuart, and Anandkumar]{kovachki2021neural}
Nikola~B. Kovachki, Zongyi Li, Burigede Liu, Kamyar Azizzadenesheli, Kaushik
  Bhattacharya, Andrew~M. Stuart, and Anima Anandkumar.
\newblock Neural operator: Learning maps between function spaces.
\newblock \emph{CoRR}, abs/2108.08481, 2021.

\bibitem[Kreis(2002)]{kreis_frequency_2002}
Thomas~M. Kreis.
\newblock Frequency analysis of digital holography.
\newblock \emph{Optical Engineering}, 41\penalty0 (4):\penalty0 771, 2002.

\bibitem[Li et~al.(2020)Li, Kovachki, Azizzadenesheli, Liu, Bhattacharya,
  Stuart, and Anandkumar]{DBLP:journals/corr/abs-2010-08895}
Zongyi Li, Nikola~B. Kovachki, Kamyar Azizzadenesheli, Burigede Liu, Kaushik
  Bhattacharya, Andrew~M. Stuart, and Anima Anandkumar.
\newblock Fourier neural operator for parametric partial differential
  equations.
\newblock \emph{CoRR}, abs/2010.08895, 2020.

\bibitem[Liu et~al.(2022)Liu, Mao, Wu, Feichtenhofer, Darrell, and
  Xie]{liu2022convnet2020s}
Zhuang Liu, Hanzi Mao, Chao-Yuan Wu, Christoph Feichtenhofer, Trevor Darrell,
  and Saining Xie.
\newblock A convnet for the 2020s, 2022.

\bibitem[Mallery and Hong(2019)]{Mallery:19}
Kevin Mallery and Jiarong Hong.
\newblock Regularized inverse holographic volume reconstruction for 3d particle
  tracking.
\newblock \emph{Opt. Express}, 27\penalty0 (13):\penalty0 18069--18084, 2019.

\bibitem[Needell and Tropp(2009)]{NEEDELL2009301}
D. Needell and J.A. Tropp.
\newblock Cosamp: Iterative signal recovery from incomplete and inaccurate
  samples.
\newblock \emph{Applied and Computational Harmonic Analysis}, 26\penalty0
  (3):\penalty0 301--321, 2009.

\bibitem[Raonić et~al.(2023)Raonić, Molinaro, Ryck, Rohner, Bartolucci,
  Alaifari, Mishra, and de~Bézenac]{raonić2023convolutional}
Bogdan Raonić, Roberto Molinaro, Tim~De Ryck, Tobias Rohner, Francesca
  Bartolucci, Rima Alaifari, Siddhartha Mishra, and Emmanuel de Bézenac.
\newblock Convolutional neural operators for robust and accurate learning of
  pdes, 2023.

\bibitem[Rayleigh(1879)]{doi:10.1080/14786447908639684Rayleigh}
Rayleigh.
\newblock Xxxi. investigations in optics, with special reference to the
  spectroscope.
\newblock \emph{The London, Edinburgh, and Dublin Philosophical Magazine and
  Journal of Science}, 8\penalty0 (49):\penalty0 261--274, 1879.

\bibitem[Ronneberger et~al.(2015)Ronneberger, Fischer, and
  Brox]{DBLP:journals/corr/RonnebergerFB15}
Olaf Ronneberger, Philipp Fischer, and Thomas Brox.
\newblock U-net: Convolutional networks for biomedical image segmentation.
\newblock \emph{CoRR}, abs/1505.04597, 2015.

\bibitem[Schlenczek(2018)]{schlenczek2018airborne}
Oliver Schlenczek.
\newblock \emph{Airborne and ground-based holographic measurement of
  hydrometeors in liquid-phase, mixed-phase and ice clouds}.
\newblock PhD thesis, University of Mainz, 2018.

\bibitem[Schreck et~al.(2022)Schreck, Gantos, Hayman, Bansemer, and
  Gagne]{Schreck}
J.~S. Schreck, G. Gantos, M. Hayman, A. Bansemer, and D.~J. Gagne.
\newblock Neural network processing of holographic images.
\newblock \emph{Atmos. Meas. Tech.}, 15:\penalty0 5793–5819, 2022.

\bibitem[Seifi et~al.(2012)Seifi, Fournier, Denis, Chareyron, and
  Marié]{seifi_three-dimensional_2012}
Mozhdeh Seifi, Corinne Fournier, Loic Denis, Delphine Chareyron, and Jean-Louis
  Marié.
\newblock Three-dimensional reconstruction of particle holograms: a fast and
  accurate multiscale approach.
\newblock \emph{Journal of the Optical Society of America A}, 29\penalty0
  (9):\penalty0 1808, 2012.

\bibitem[Shao et~al.(2020)Shao, Mallery, Kumar, and Hong]{Shao:20}
Siyao Shao, Kevin Mallery, S.~Santosh Kumar, and Jiarong Hong.
\newblock Machine learning holography for 3d particle field imaging.
\newblock \emph{Opt. Express}, 28\penalty0 (3):\penalty0 2987--2999, 2020.

\bibitem[Shimobaba et~al.(2019)Shimobaba, Takahashi, Yamamoto, Endo, Shiraki,
  Nishitsuji, Hoshikawa, Kakue, and Ito]{Shimobaba:19}
Tomoyoshi Shimobaba, Takayuki Takahashi, Yota Yamamoto, Yutaka Endo, Atsushi
  Shiraki, Takashi Nishitsuji, Naoto Hoshikawa, Takashi Kakue, and Tomoyosh
  Ito.
\newblock Digital holographic particle volume reconstruction using a deep
  neural network.
\newblock \emph{Appl. Opt.}, 58\penalty0 (8):\penalty0 1900--1906, 2019.

\bibitem[Sotthivirat and Fessler(2004)]{sotthivirat2004penalized}
Sayan Sotthivirat and Jeffrey~A Fessler.
\newblock Penalized-likelihood image reconstruction for digital holography.
\newblock \emph{Journal of the Optical Society of America A}, 21\penalty0
  (5):\penalty0 737, 2004.

\bibitem[Soulez et~al.(2007)Soulez, Denis, Fournier, \'{E}ric Thi\'{e}baut, and
  Goepfert]{Soulez:07}
Ferr\'{e}ol Soulez, Lo\"{i}c Denis, Corinne Fournier, \'{E}ric Thi\'{e}baut,
  and Charles Goepfert.
\newblock Inverse-problem approach for particle digital holography: accurate
  location based on local optimization.
\newblock \emph{J. Opt. Soc. Am. A}, 24\penalty0 (4):\penalty0 1164--1171,
  2007.

\bibitem[Stevens et~al.(2021)Stevens, Bony, Farrell, Ament, Blyth, Fairall,
  Karstensen, Quinn, Speich, Acquistapace, Aemisegger, Albright, Bellenger,
  Bodenschatz, Caesar, Chewitt-Lucas, de~Boer, Delano\"e, Denby, Ewald,
  Fildier, Forde, George, Gross, Hagen, Hausold, Heywood, Hirsch, Jacob,
  Jansen, Kinne, Klocke, K\"olling, Konow, Lothon, Mohr, Naumann, Nuijens,
  Olivier, Pincus, P\"ohlker, Reverdin, Roberts, Schnitt, Schulz, Siebesma,
  Stephan, Sullivan, Touz\'e-Peiffer, Vial, Vogel, Zuidema, Alexander, Alves,
  Arixi, Asmath, Bagheri, Baier, Bailey, Baranowski, Baron, Barrau, Barrett,
  Batier, Behrendt, Bendinger, Beucher, Bigorre, Blades, Blossey, Bock,
  B\"oing, Bosser, Bourras, Bouruet-Aubertot, Bower, Branellec, Branger,
  Brennek, Brewer, Brilouet, Br\"ugmann, Buehler, Burke, Burton, Calmer,
  Canonici, Carton, Cato~Jr., Charles, Chazette, Chen, Chilinski, Choularton,
  Chuang, Clarke, Coe, Cornet, Coutris, Couvreux, Crewell, Cronin, Cui,
  Cuypers, Daley, Damerell, Dauhut, Deneke, Desbios, D\"orner, Donner, Douet,
  Drushka, D\"utsch, Ehrlich, Emanuel, Emmanouilidis, Etienne, Etienne-Leblanc,
  Faure, Feingold, Ferrero, Fix, Flamant, Flatau, Foltz, Forster, Furtuna,
  Gadian, Galewsky, Gallagher, Gallimore, Gaston, Gentemann, Geyskens, Giez,
  Gollop, Gouirand, Gourbeyre, de~Graaf, de~Groot, Grosz, G\"uttler, Gutleben,
  Hall, Harris, Helfer, Henze, Herbert, Holanda, Ibanez-Landeta, Intrieri,
  Iyer, Julien, Kalesse, Kazil, Kellman, Kidane, Kirchner, Klingebiel,
  K\"orner, Kremper, Kretzschmar, Kr\"uger, Kumala, Kurz, L'H\'egaret, Labaste,
  Lachlan-Cope, Laing, Landsch\"utzer, Lang, Lange, Lange, Laplace, Lavik,
  Laxenaire, Le~Bihan, Leandro, Lefevre, Lena, Lenschow, Li, Lloyd, Los, Losi,
  Lovell, Luneau, Makuch, Malinowski, Manta, Marinou, Marsden, Masson, Maury,
  Mayer, Mayers-Als, Mazel, McGeary, McWilliams, Mech, Mehlmann, Meroni,
  Mieslinger, Minikin, Minnett, M\"oller, Morfa~Avalos, Muller, Musat, Napoli,
  Neuberger, Noisel, Noone, Nordsiek, Nowak, Oswald, Parker, Peck, Person,
  Philippi, Plueddemann, P\"ohlker, P\"ortge, P\"oschl, Pologne, Posyniak,
  Prange, Qui\~nones Mel\'endez, Radtke, Ramage, Reimann, Renault, Reus, Reyes,
  Ribbe, Ringel, Ritschel, Rocha, Rochetin, R\"ottenbacher, Rollo, Royer,
  Sadoulet, Saffin, Sandiford, Sandu, Sch\"afer, Schemann, Schirmacher,
  Schlenczek, Schmidt, Schr\"oder, Schwarzenboeck, Sealy, Senff, Serikov,
  Shohan, Siddle, Smirnov, Sp\"ath, Spooner, Stolla, Szk\'o{\l}ka, de~Szoeke,
  Tarot, Tetoni, Thompson, Thomson, Tomassini, Totems, Ubele, Villiger, von
  Arx, Wagner, Walther, Webber, Wendisch, Whitehall, Wiltshire, Wing, Wirth,
  Wiskandt, Wolf, Worbes, Wright, Wulfmeyer, Young, Zhang, Zhang, Ziemen,
  Zinner, and Z\"oger]{essd-13-4067-2021}
B. Stevens, S. Bony, D. Farrell, F. Ament, A. Blyth, C. Fairall, J. Karstensen,
  P.~K. Quinn, S. Speich, C. Acquistapace, F. Aemisegger, A.~L. Albright, H.
  Bellenger, E. Bodenschatz, K.-A. Caesar, R. Chewitt-Lucas, G. de Boer, J.
  Delano\"e, L. Denby, F. Ewald, B. Fildier, M. Forde, G. George, S. Gross, M.
  Hagen, A. Hausold, K.~J. Heywood, L. Hirsch, M. Jacob, F. Jansen, S. Kinne,
  D. Klocke, T. K\"olling, H. Konow, M. Lothon, W. Mohr, A.~K. Naumann, L.
  Nuijens, L. Olivier, R. Pincus, M. P\"ohlker, G. Reverdin, G. Roberts, S.
  Schnitt, H. Schulz, A.~P. Siebesma, C.~C. Stephan, P. Sullivan, L.
  Touz\'e-Peiffer, J. Vial, R. Vogel, P. Zuidema, N. Alexander, L. Alves, S.
  Arixi, H. Asmath, G. Bagheri, K. Baier, A. Bailey, D. Baranowski, A. Baron,
  S. Barrau, P.~A. Barrett, F. Batier, A. Behrendt, A. Bendinger, F. Beucher,
  S. Bigorre, E. Blades, P. Blossey, O. Bock, S. B\"oing, P. Bosser, D.
  Bourras, P. Bouruet-Aubertot, K. Bower, P. Branellec, H. Branger, M. Brennek,
  A. Brewer, P.-E. Brilouet, B. Br\"ugmann, S.~A. Buehler, E. Burke, R. Burton,
  R. Calmer, J.-C. Canonici, X. Carton, G. Cato~Jr., J.~A. Charles, P.
  Chazette, Y. Chen, M.~T. Chilinski, T. Choularton, P. Chuang, S. Clarke, H.
  Coe, C. Cornet, P. Coutris, F. Couvreux, S. Crewell, T. Cronin, Z. Cui, Y.
  Cuypers, A. Daley, G.~M. Damerell, T. Dauhut, H. Deneke, J.-P. Desbios, S.
  D\"orner, S. Donner, V. Douet, K. Drushka, M. D\"utsch, A. Ehrlich, K.
  Emanuel, A. Emmanouilidis, J.-C. Etienne, S. Etienne-Leblanc, G. Faure, G.
  Feingold, L. Ferrero, A. Fix, C. Flamant, P.~J. Flatau, G.~R. Foltz, L.
  Forster, I. Furtuna, A. Gadian, J. Galewsky, M. Gallagher, P. Gallimore, C.
  Gaston, C. Gentemann, N. Geyskens, A. Giez, J. Gollop, I. Gouirand, C.
  Gourbeyre, D. de Graaf, G.~E. de Groot, R. Grosz, J. G\"uttler, M. Gutleben,
  K. Hall, G. Harris, K.~C. Helfer, D. Henze, C. Herbert, B. Holanda, A.
  Ibanez-Landeta, J. Intrieri, S. Iyer, F. Julien, H. Kalesse, J. Kazil, A.
  Kellman, A.~T. Kidane, U. Kirchner, M. Klingebiel, M. K\"orner, L.~A.
  Kremper, J. Kretzschmar, O. Kr\"uger, W. Kumala, A. Kurz, P. L'H\'egaret, M.
  Labaste, T. Lachlan-Cope, A. Laing, P. Landsch\"utzer, T. Lang, D. Lange, I.
  Lange, C. Laplace, G. Lavik, R. Laxenaire, C. Le~Bihan, M. Leandro, N.
  Lefevre, M. Lena, D. Lenschow, Q. Li, G. Lloyd, S. Los, N. Losi, O. Lovell,
  C. Luneau, P. Makuch, S. Malinowski, G. Manta, E. Marinou, N. Marsden, S.
  Masson, N. Maury, B. Mayer, M. Mayers-Als, C. Mazel, W. McGeary, J.~C.
  McWilliams, M. Mech, M. Mehlmann, A.~N. Meroni, T. Mieslinger, A. Minikin, P.
  Minnett, G. M\"oller, Y. Morfa~Avalos, C. Muller, I. Musat, A. Napoli, A.
  Neuberger, C. Noisel, D. Noone, F. Nordsiek, J.~L. Nowak, L. Oswald, D.~J.
  Parker, C. Peck, R. Person, M. Philippi, A. Plueddemann, C. P\"ohlker, V.
  P\"ortge, U. P\"oschl, L. Pologne, M. Posyniak, M. Prange, E. Qui\~nones
  Mel\'endez, J. Radtke, K. Ramage, J. Reimann, L. Renault, K. Reus, A. Reyes,
  J. Ribbe, M. Ringel, M. Ritschel, C.~B. Rocha, N. Rochetin, J.
  R\"ottenbacher, C. Rollo, H. Royer, P. Sadoulet, L. Saffin, S. Sandiford, I.
  Sandu, M. Sch\"afer, V. Schemann, I. Schirmacher, O. Schlenczek, J. Schmidt,
  M. Schr\"oder, A. Schwarzenboeck, A. Sealy, C.~J. Senff, I. Serikov, S.
  Shohan, E. Siddle, A. Smirnov, F. Sp\"ath, B. Spooner, M.~K. Stolla, W.
  Szk\'o{\l}ka, S.~P. de Szoeke, S. Tarot, E. Tetoni, E. Thompson, J. Thomson,
  L. Tomassini, J. Totems, A.~A. Ubele, L. Villiger, J. von Arx, T. Wagner, A.
  Walther, B. Webber, M. Wendisch, S. Whitehall, A. Wiltshire, A.~A. Wing, M.
  Wirth, J. Wiskandt, K. Wolf, L. Worbes, E. Wright, V. Wulfmeyer, S. Young, C.
  Zhang, D. Zhang, F. Ziemen, T. Zinner, and M. Z\"oger.
\newblock Eurec$^4$a.
\newblock \emph{Earth System Science Data}, 13\penalty0 (8):\penalty0
  4067--4119, 2021.

\bibitem[Thiede et~al.(2025)Thiede, Schlenczek, Stieger, Ecker, Bodenschatz,
  and Bagheri]{methodspaper}
B. Thiede, O. Schlenczek, K. Stieger, A. Ecker, E. Bodenschatz, and
  Gholamhossein Bagheri.
\newblock In-line holographic droplet imaging: Accelerated classification with
  convolutional neural networks and quantitative experimental validation.
\newblock \emph{Atmospheric Measurement Techniques Preprint}, 2025.

\bibitem[Thompson(1974)]{thompson1974}
B.~J. Thompson.
\newblock Holographic particle sizing techniques.
\newblock \emph{Journal of Physics E: Scientific Instruments}, 7:\penalty0 781,
  1974.

\bibitem[Touloupas et~al.(2020)Touloupas, Lauber, Henneberger, Beck, and
  Lucchi]{amt-13-2219-2020}
G. Touloupas, A. Lauber, J. Henneberger, A. Beck, and A. Lucchi.
\newblock A convolutional neural network for classifying cloud particles
  recorded by imaging probes.
\newblock \emph{Atmospheric Measurement Techniques}, 13\penalty0 (5):\penalty0
  2219--2239, 2020.

\bibitem[Wang et~al.(2020)Wang, Wu, Huang, and Xing]{wang2020high}
Haohan Wang, Xindi Wu, Zeyi Huang, and Eric~P. Xing.
\newblock High frequency component helps explain the generalization of
  convolutional neural networks, 2020.

\bibitem[Wu et~al.(2024{\natexlab{a}})Wu, Wang, Thoroddsen, and
  Chen]{ni_chen_diffHolo}
Yang Wu, Jun Wang, Sigurdur~T. Thoroddsen, and Ni Chen.
\newblock Single-shot high-density volumetric particle imaging enabled by
  differentiable holography.
\newblock \emph{IEEE Transactions on Industrial Informatics}, 20\penalty0
  (12):\penalty0 13696--13706, 2024{\natexlab{a}}.

\bibitem[Wu et~al.(2024{\natexlab{b}})Wu, Wang, Thoroddsen, and
  Chen]{partial3H}
Yang Wu, Jun Wang, Sigurdur~T. Thoroddsen, and Ni Chen.
\newblock Single-shot high-density volumetric particle imaging enabled by
  differentiable holography.
\newblock \emph{IEEE Transactions on Industrial Informatics}, 20\penalty0
  (12):\penalty0 13696--13706, 2024{\natexlab{b}}.

\bibitem[Xu et~al.(2024)Xu, Liu, and
  Zhang]{xu2024dilatedconvolutionneuraloperator}
Bo Xu, Xinliang Liu, and Lei Zhang.
\newblock Dilated convolution neural operator for multiscale partial
  differential equations, 2024.

\bibitem[Zhang et~al.(2022)Zhang, Zhu, and Lam]{Zhang:22}
Yunping Zhang, Yanmin Zhu, and Edmund~Y. Lam.
\newblock Holographic 3d particle reconstruction using a one-stage network.
\newblock \emph{Appl. Opt.}, 61\penalty0 (5):\penalty0 B111--B120, 2022.

\end{thebibliography}
}

\clearpage
\appendix
\setcounter{page}{1}
\maketitlesupplementary
\counterwithin*{equation}{section}
\renewcommand\theequation{\thesection\arabic{equation}}

\section{Analyzing the Power Spectrum}
\label{appendix:visual_power_spectrum}
\begin{figure}[h]
  \centering
   \includegraphics[width=0.7\linewidth]{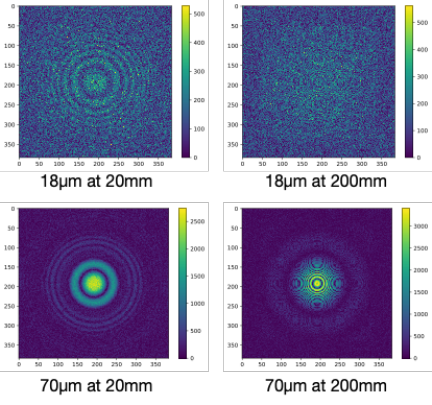}
   \caption{Power spectrum for large and small particles at near and far depths to the camera. Power spectrum of near particles (left) can be seen as a zoomed-in version of farther particles (right).}
   \label{fig:pow_spec}
\end{figure}

\section{On Choosing $\lambda_w$ and $d_w$}
\label{appendix:lam_w_hyperparas}
Depending on the particle density one wants to image and the requirement of lightweighted-ness of the particle detection step, these hyperparameters can be tuned. We aimed to reduce the overlap of fringes in the weighted hologram $\mathcal{H}_w$ by localizing them, such that a lightweight detector (a moderately shallow U-net) with relatively small receptive field suffices to capture the de-mixed fringe patterns. With a much longer wavelength ($\lambda /4$), the fringe patterns are still too large. On the other hand, a much shorter wavelength ($\lambda /16$) turns every particle into mostly a dot that almost looks like the final target of the particle detection step, but the CSFM network alone could not solve the full detection task well. Hence the two-step procedure. It is possible that further tuning the target wavelength within $\pm$50\% would improve performance somewhat, but we did not investigate this yet.

\begin{figure}[h!]
    \centering
    \includegraphics[width=\linewidth]{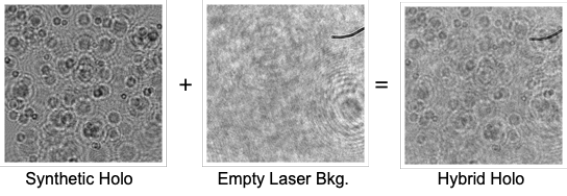}
    \caption{Example of hybrid hologram crop (\fig{fig:hybrid_holo_ex}). Empty laser background (bkg.) in \fig{fig:hybrid_holo_ex} is a crop ($1536\times 1536$) from $9344\times7000$ background with 3.2\textmu m pixel pitch and recorded at 532\,nm. }
    \label{fig:hybrid_holo_ex}
\end{figure}

\section{Architecture and Training Details}
\paragraph{U-net.} The input hologram is first lifted to N\textsubscript{L} channels, and consequently processed through a series of downsampling convolutional blocks -- the encoder section -- and residual convolutional skip connections. 
\begin{figure}[h!]
    \centering
    \includegraphics[width=\linewidth]{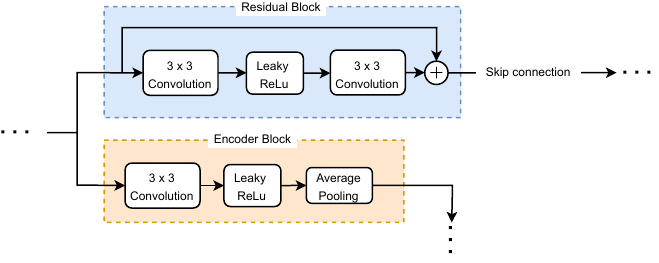}
    \caption{Schemata on the architecture of a U-Net encoder layer.}
    \label{fig:unet-encoder-arch}
\end{figure}
Each encoder layer, represented in \fig{fig:unet-encoder-arch}, is composed of a downsampling convolutional block, termed encoder block, and n\textsubscript{res} residual convolutional blocks that comprise the processing assigned to each skip connection. The encoder block applies a single $3\times3$ convolution, followed by a Leaky ReLu as a nonlinearity, and an average pooling operation. On the other hand, the skip connection-associated residual convolutional blocks apply two $3\times3$ convolutions consecutively, interspersed by a leaky ReLU as a non-linearity, concatenating its input with the output. Note that for a five layer deep U-net there will be four skip connections. The encoder section is connected to the decoder through the bottleneck, which is composed of a series of n\textsubscript{res\textunderscore neck} residual convolutional blocks. 

Similarly to the encoder, the decoder section applies a series of upsampling convolutional blocks.

\begin{figure}[h!]
    \centering
    \includegraphics[width=\linewidth]{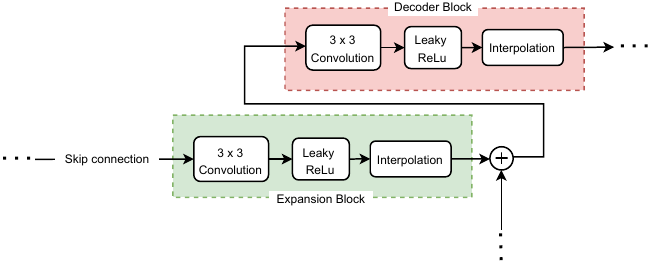}
    \caption{Schemata on the architecture of a U-Net decoder layer.}
    \label{fig:unet-decoder-arch}
\end{figure}
Within each decoder layer, visualized in \fig{fig:unet-decoder-arch}, the output of the respective skip layer is processed by a convolutional upsampling block - \emph{expansion block} - before being concatenated with the output of the last layer and passed into a second convolutional upsampling block - \emph{decoder block}. Both of these block are composed of a $3\times3$ convolution, followed by a leaky ReLu as a nonlinearity and a bilinear interpolation. These blocks are repeated n$_\text{layers}$ times, reaching the original input size with N$_\text{P}$ channels. A final residual convolutional block is placed with a sigmoid in place of leaky ReLU to limit the range between $[0,1]$, outputting 3 channels - $xy,z,d$. For $z$ and $d$, the channels are linearly scaled to 200\,mm and 100\,µm, respectively. 

\paragraph{Hyperparameters.}\tab{tab:hyperparams} details the final hyperparameters used for training FLASHµ. 
\label{appendix:Training Details and Hyperparameters}
\begin{table}[h]
    \centering
    \footnotesize
    \begin{tabular}{ll}
        \toprule
        \textbf{Hyperparameter} & \textbf{Value} \\
        \midrule
        \multicolumn{2}{l}{\textbf{Dilated Fourier (DF) Part}} \\
        N\textsubscript{in} & 3 \textcolor{gray}{\# Hologram, X, Y}\\
        N\textsubscript{hidden} & 128 \\
        N\textsubscript{out} & 1 \\
        N\textsubscript{L} & 128 \textcolor{gray}{\# lifting channels} \\
        N\textsubscript{P} & 128 \textcolor{gray}{\# projection channels}\\
        n\textsubscript{modes} & 384\\ 
        n\textsubscript{layers} &  3 \textcolor{gray}{\# 1 DF-block is always added. Total = 4}\\
        separability & [False, True, True, True] \textcolor{gray}{\# 4 DF blocks}\\
        dilation & 2  \textcolor{gray}{\# 1 means no dilation/masking}\\
        interpolation & True \\
        skip & ``conv3'' \textcolor{gray}{\# 3x3 convolution}\\
        factorization & ``cp''  \textcolor{gray}{\# `tucker', `dense'}\\
        implementation & ``factorized'' \\
        rank & 16 \textcolor{gray}{\# Kruskal rank. Can put float too.}\\
        bias & True \\
        batchnorm & False \\
        activations & ``gelu''\\
        \midrule
        \multicolumn{2}{l}{\textbf{Dilated Convolutional (DC) Part}} \\
        k\textsubscript{size} & 5 \\
        dilations & [1, 3, 9] \\
        padding & True \\
        activations & ``lrelu'' \\
        \midrule \midrule
        \multicolumn{2}{l}{\textbf{U-Net Part}} \\
        N\textsubscript{in} & 2 \textcolor{gray}{\# 1 for vanilla U-net baseline}\\
        N\textsubscript{out} & 3 \\
        N\textsubscript{L} & 64 \textcolor{gray}{\# lifting channels} \\
        N\textsubscript{P} & 64 \textcolor{gray}{\# projection channels} \\
        n\textsubscript{in} & 384 \\
        n\textsubscript{layers} & 5 \\
        n\textsubscript{res} & 2 \\
        activations & ``lrelu'' \\
        \bottomrule
    \end{tabular}
    \caption{Hyperparameters for FLASHµ. Both DF and DC refer to the CSFM, while the U-Net part is related to its namesake.}
    \label{tab:hyperparams}
\end{table}
\paragraph{Training Details and Data Augmentation}
Input holograms are standardized using the mean and standard deviation of the training data. During training, samples are randomly flipped vertically, horizontally and rotated $\in\{0,90,180,270\}$. Additionally, the brightness of a sample is randomly jittered between $\pm 0-10\,\%$.

At first, the two stages of FLASHµ are trained separately. The first stage, CSFM, of the network with $k=384//2$ took approximately three and half days on a single NVIDIA 40\,GB A100 GPU, with a batch size of 12 using the Adam optimizer and an initial learning rate of $5\times10^{-4}$, which was reduced by a factor of 0.9 if the validation loss did not decrease for three epochs. Training stopped when the loss remained plateaued for five epochs. The U-net is trained with the predicted weighted holograms from the trained FLASHµ and concatenated with input hologram. This stage took approximately 2 days with batch size of 64 and learning rate of $1\times10^{-4}$. LR Scheduler and early stopping were set in the same fashion as CSFM.

\section{Ablations}
We provide ablations of hyperparameters which we found had major effect on performance. Training data for the ablations below is described in section \ref{sec:exp}. Metrics tables 1-3 and 5-9 are evaluated on CloudTarget using a single U-net detection head trained on ground-truth weighted holograms to be fair for all set of hyperparameters.  
For \tab{tab:ablation_crop_sz}, CSFM was fixed with n$_\text{modes}$=64 for all crop sizes, no dilation of Fourier weight, no separable spectral convolutions and tensor decompositions. The rest of the hyperparameters are mentioned in \tab{tab:hyperparams}. \tab{tab:ablation_ker_sz} was done with fixed crop size of 384 and varying Fourier kernel sizes n$_\text{modes}$, without separability and CP decomposition. n$_\text{modes}$//2 signifies that n$_\text{modes}$ were covered in each direction, but half of the modes were masked and then interpolated. \tab{tab:ablation_cp_sep} introduced spectral separability and CP decomposition for the former models. 
In the following, certain ablations are not done for all crop sizes because training at least takes 2-3 days per hyperparameter configuration, but the trend is clear from smaller crop sizes and we expect it to generalize to larger ones. 
In \tab{tab:ablation_training_samples}, Tucker decomposition was used in place of CP -- this was done in the earlier stages of the project. Precision, Recall and F1 score was only done in $xy$ for ablations, relaxing the 10\,mm $z$ constraint. 
\label{appendix:ablations}
\begin{table}[h]
    \centering
    \begin{tabular}{|c|c|c|c|c|c|c|c|}
        \#Samples & Prec. & Rec. & Peak F1 \\
        \hline
        54000  & 0.74 & 0.63 & 0.68  \\
        108000 & 0.73 & 0.68 & 0.70 \\
        216000 & 0.72 & 0.70 & 0.71 
    \end{tabular}
    \caption{Effect of number of training samples. Architecture is fixed (n$_\text{modes}=256$, n$_\text{layers} = 3$, dilation = 2, factorization = `tucker', rank = 0.51, rest is \tab{tab:hyperparams}). Crop size is 256. Metrics are evaluated on CloudTarget.}
    \label{tab:ablation_training_samples}
\end{table}

\begin{table}[h]
    \centering
    \begin{tabular}{c|c|c|c|c|c|c}
        Crop Sz.  & DC & P & R & F1 \\
        \hline 
        128 & N  & 0.753 & 0.419 & 0.539 \\
        128 & Y  & 0.731 & 0.482 & 0.581 \\
        256 & N  & 0.731 & 0.544 & 0.624 \\
        256 & Y  & 0.722 & 0.625 & 0.670 
    \end{tabular}
    \caption{Effect of adding dilated convolution (DC) blocks on performance. Also the effect of increasing crop size  and Fourier kernel size is apparent. Y - Yes,  N - No. n$_\text{modes}$ = 64 and 128 for crop size 128 and 256 respectively. Dilation = 1 (no dilation). No separable spectral convolutions and no tensor decomposition. n$_\text{layers}$=3. Rest follows from \tab{tab:hyperparams}.} 
    \label{tab:ablation_add_dc}
\end{table}

\begin{table}[h]
    \centering
    \begin{tabular}{c|c|c|c|c}
        Crop Sz. & Interp. & P & R & F1 \\
         \hline
         384 & N & 0.67 & 0.60 & 0.63\\
         384 & Y & 0.72 & 0.79 & 0.75 
    \end{tabular}
    \caption{Effect on not interpolating after masking or dilating alternating frequencies on detection metrics. Hyperparameters (from \tab{tab:hyperparams}) are fixed except the interpolation variable which is set to True and False. Y - Yes,  N - No.}
    \label{tab:ablation_interp}
\end{table}

\begin{table}[h]
    \centering
    \begin{tabular}{c|c|c|c|c}
         Crop Sz. &n$_\text{layers}$ & P & R & F1 \\
         \hline
         384 & 2 & 0.64 & 0.43 & 0.51 \\
         384 & 3 & 0.71 & 0.80 & 0.75 \\
         384 & 4 & 0.68 & 0.82 & 0.75
    \end{tabular}
    \caption{Ablation for number of DF-DC blocks (n$_\text{layers}$) in CSFM. Hyperparameters (from \tab{tab:hyperparams}) are fixed, except n$_\text{layers}=2,3,\, \text{and}\,4$. n$_\text{layers}$=3 would mean 4 DF blocks and 3 DC blocks. }
    \label{tab:ablation_n_layers}
\end{table}

\begin{table}[h]
    \centering
    \begin{tabular}{c|c|c|c|c}
         Crop Sz & $N_\text{hidden}$ & P & R & F1 \\
        \hline 
        384 & 64 & 0.69 & 0.63 & 0.66\\ 
        384 & 96 & 0.72 & 0.73 & 0.72\\
        384 & 128 & 0.71 & 0.80 & 0.75
    \end{tabular}
    \caption{Ablation for width or number of hidden channels of the CSFM network. Hyperparameters (from \tab{tab:hyperparams}) are fixed, except N$_\text{hidden}$.}
    \label{tab:ablation_width}
\end{table}

\section{Further Analysis on CloudTarget}
\label{appendix:error_size_z_hist}
\begin{figure*}[t]
  \centering
    \includegraphics[width = \linewidth]{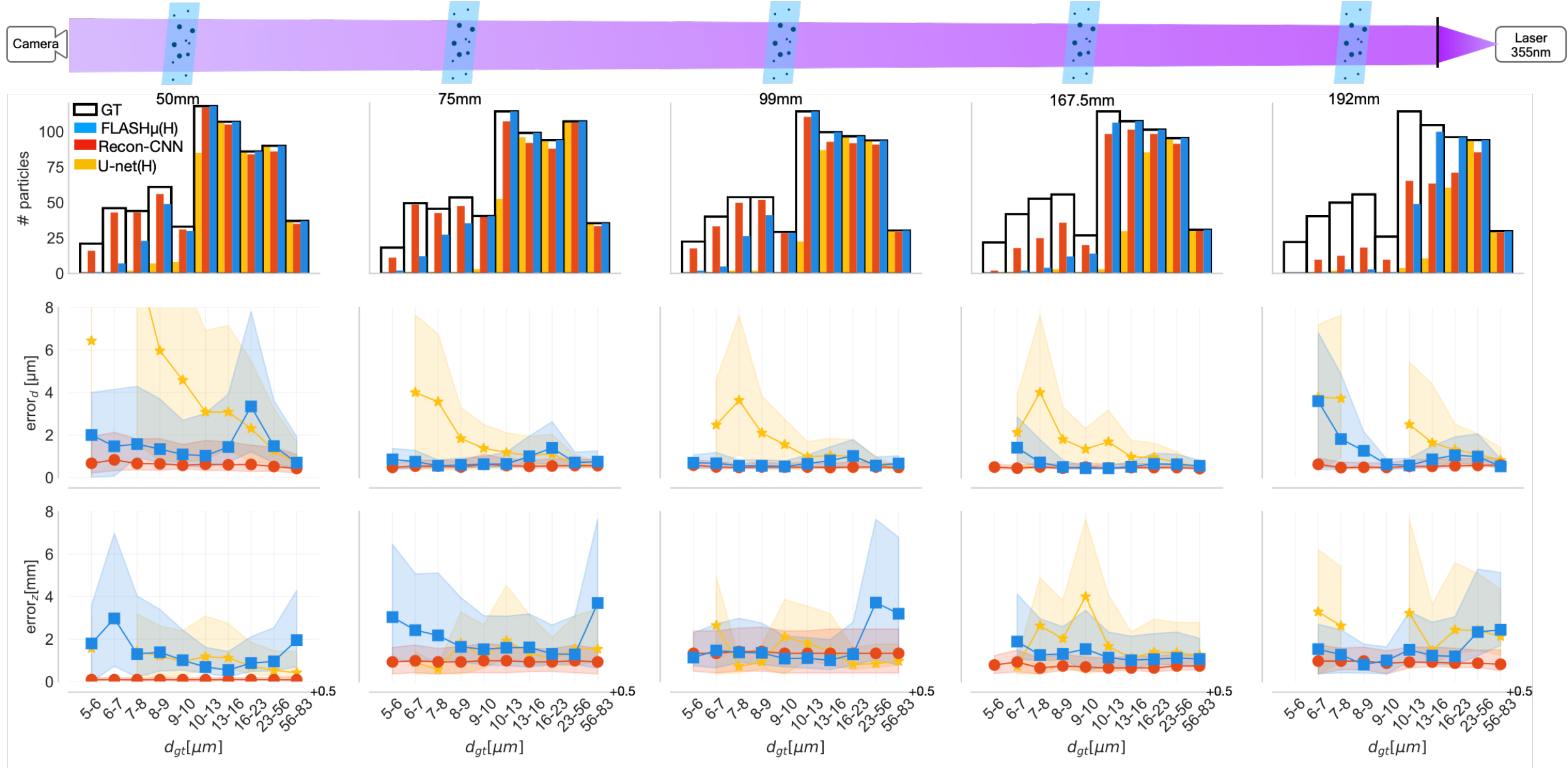}
    \caption{Analysis on CloudTarget. CloudTarget was moved along $z$ and shaken in $xy$. 
    Row 1 (top row) plots the histogram of particles detected for each size bin. Row 2 (middle row) shows the median error of the absolute size ($d$) error and interquartile range per size bin.
    Row 3 (bottom row) shows the median depth ($z$) error with interquartile range per size bin. $d_{gt}$ stands for ground truth size (diameter).
    }
    \label{fig:error_plots}
\end{figure*}

 \fig{fig:error_plots} further details the quality of size and depth regression of FLASHµ per $z$ slice and size bins on CloudTarget. 

\section{Cloud Holograms}
\label{appendix:cloud_holograms}
\begin{figure*}[t]
  \centering
  \begin{subfigure}{\linewidth}
    \includegraphics[width = \linewidth]{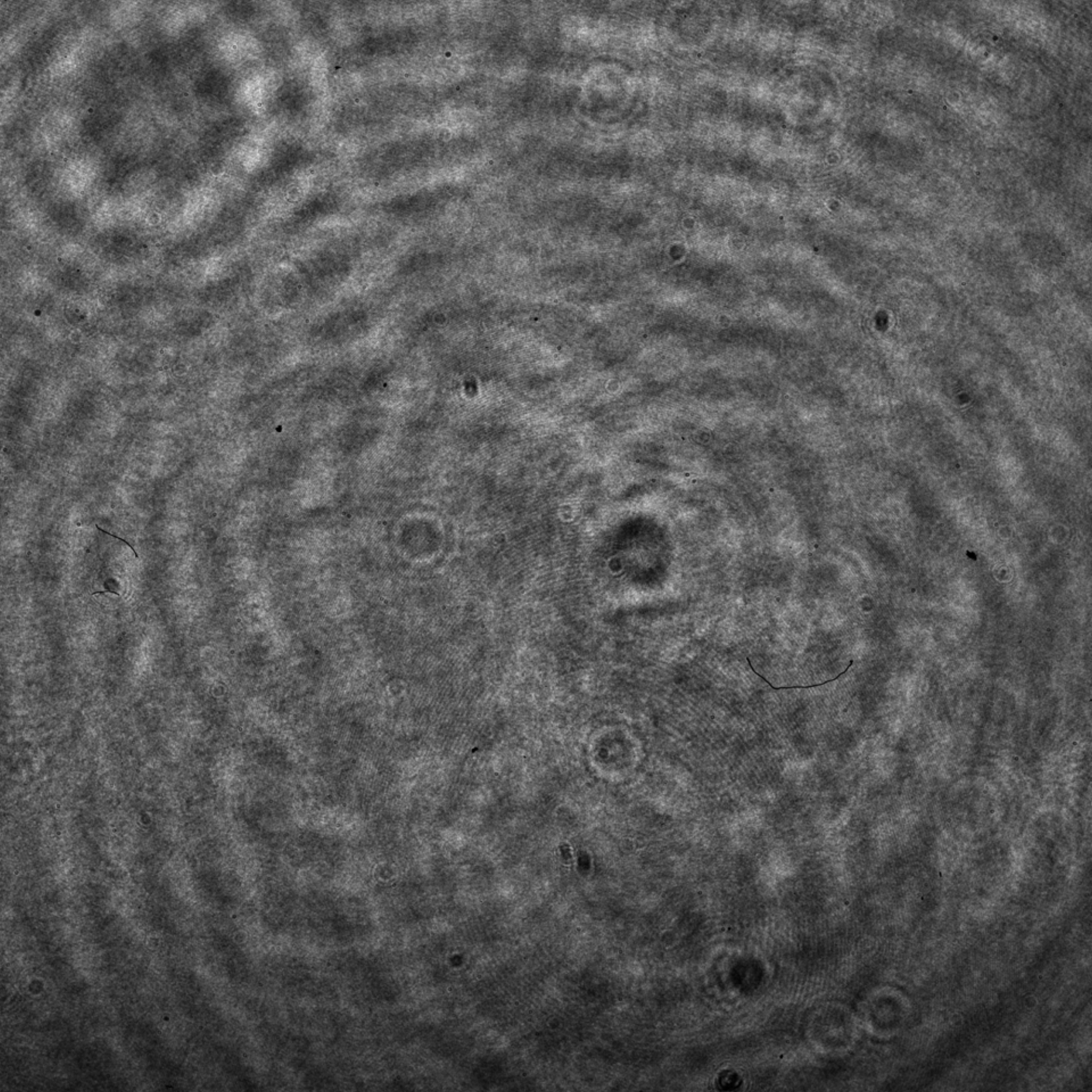}
    
  \end{subfigure}
  \caption{An example of a real cloud hologram (5120$\times 5120$ pixels).
    }
    \label{fig:real_cloud}
\end{figure*}

\begin{figure*}[t]
  \centering
  \begin{subfigure}{\linewidth}
    \includegraphics[width = \linewidth]{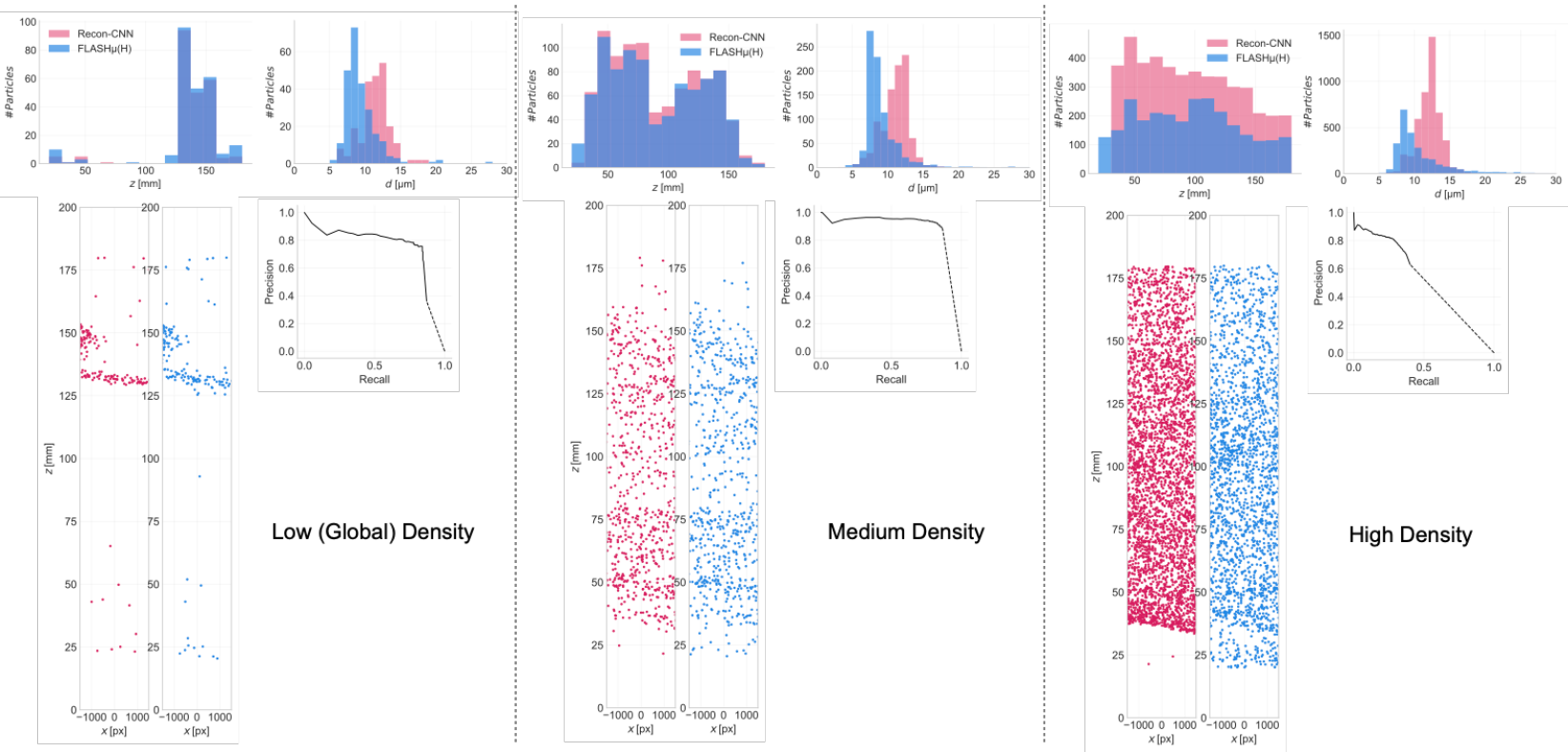}
  \end{subfigure}
  \caption{Evaluation of FLASHµ on real cloud holograms of low, medium and high global density in comparison to Recon-CNN treated as ground truth. In each column, first row contains the predicted histograms from the two methods - depth ($z$) on the left and size (diameter $d$) on the right. Row 2 contains a visual of the ($x,z$) predictions (left) and Precision-Recall is plotted of FLASHµ is plotted with respect to Recon-CNN predictions. 
    }
    \label{fig:low_med_high}
\end{figure*}

\fig{fig:real_cloud} is an example of a real cloud hologram shown for better understanding of severity of noise.  
In the following \fig{fig:low_med_high}, we show evaluation of FLASHµ on three cloud holograms - low, medium and highly dense clouds, against Recon-CNN. The two methods seem to get similar spatial distribution $(x,y,z)$ for low and medium dense holograms and appreciable match in terms precision and recall of 80\%-90\% in them. Size histograms don't seem to match as much as $z$ histograms. Note that particles are majoritively small - (6-15\,µm or 2-5\,px), and Recon-CNN counts pixels for sizing -- hence the plausible mismatch. Yet it is imperative that the two methods agree on the sizing, and therefore sizing needs more scrutiny in future work. 
FLASHµ is also limited by resolution and hence has an upper bound on the number density where it can perform reliably which is yet to explored-- but from \fig{fig:low_med_high}, up till medium density holograms (around 70 particles/cm$^3$) FLASHµ seems to work decently. For denser clouds, resolution and multiscattering effects become more relevant. We're addressing these in our future work for our method.

\end{document}